\documentclass[10pt,twocolumn,letterpaper]{article}

\usepackage{style}              
\definecolor{styleblue}{rgb}{0.21,0.49,0.74}
\usepackage[pagebackref,breaklinks,colorlinks,allcolors=styleblue]{hyperref}

\title{JointDiT: Enhancing RGB-Depth Joint Modeling with Diffusion Transformers}

\author{
    Kwon Byung-Ki$^{1,2\dagger}$ \quad
    Qi Dai$^2$ \quad
    Lee Hyoseok$^1$ \quad
    Chong Luo$^2$ \quad
    Tae-Hyun Oh$^3$ \\
    \vspace{0.3em} \\  
    $^1$POSTECH \quad
    $^2$Microsoft Research Asia\quad
    $^3$KAIST
}

\usepackage{comment}
\usepackage{enumitem}
\usepackage{multirow}
\usepackage{wrapfig}
\usepackage{colortbl}
\usepackage[normalem]{ulem}

\usepackage{caption}
\usepackage{graphicx}
\usepackage{booktabs}

\setlist[itemize]{align=parleft,left=0pt}

\definecolor{kwon}{rgb}{0.36, 0.54, 0.66}
\definecolor{kjs}{rgb}{0.784, 0.003, 0.313}
\definecolor{azure(colorwheel)}{rgb}{0.0, 0.5, 1.0}
\definecolor{nicegreen}{rgb}{0.0, 0.7, 0.1}
\definecolor{ashblue}{rgb}{0.36, 0.54, 0.66}
\definecolor{ashgrey}{rgb}{0.7, 0.75, 0.71}
\definecolor{applegreen}{rgb}{0.55, 0.71, 0.0}
\definecolor{jy}{rgb}{0.58, 0, 0.827}
\definecolor{cornellred}{rgb}{0.7, 0.11, 0.11}
\definecolor{darkcyan}{rgb}{0.0, 0.55, 0.55}
\definecolor{CuGray}{gray}{0.9}
\definecolor{airforceblue}{rgb}{0.36, 0.54, 0.66}
\definecolor{rev}{rgb}{0.784, 0.003, 0.313}
\definecolor{pink}{cmyk}{0, 0.7808, 0.4429, 0.1412}
\definecolor{amethyst}{rgb}{0.6, 0.4, 0.8}
\definecolor{black}{rgb}{0.0, 0.0, 0.0}
\definecolor{tb3_yellow}{rgb}{0.996, 1.0, 0.6}
\definecolor{tb3_orange}{rgb}{0.980, 0.8, 0.604}
\definecolor{tb3_red}{rgb}{0.972, 0.6, 0.6}
\definecolor{dimgray}{rgb}{0.41, 0.41, 0.41}
\definecolor{brickred}{rgb}{0.8, 0.25, 0.33}
\definecolor{bleudefrance}{rgb}{0.19, 0.55, 0.91}
\definecolor{blue(ncs)}{rgb}{0.265, 0.445, 0.765}
\definecolor{blue(ryb)}{rgb}{0.01, 0.28, 1.0}
\definecolor{cyan}{rgb}{0.0, 1.0, 1.0}
\definecolor{darkbrown}{rgb}{0.4, 0.26, 0.13}
\definecolor{brown(traditional)}{rgb}{0.59, 0.29, 0.0}
\definecolor{hyos}{rgb}{0.662, 0.482, 0.960}
\definecolor{magenta}{rgb}{0.98, 0.176, 0.815}

\newcolumntype{g}{>{\columncolor{CuGray}}c}
\newcolumntype{z}{>{\columncolor{CuGray}}l}

\renewcommand{\paragraph}[1]{\vspace{1mm}\noindent\textbf{#1.}\,\,}

\usepackage{xspace}

\makeatletter
\def\@fnsymbol#1{\ensuremath{\ifcase#1\or *\or \dagger\or \ddagger\or
   \mathsection\or \mathparagraph\or \|\or **\or \dagger\dagger
   \or \ddagger\ddagger \else\@ctrerr\fi}}
\makeatother

\def\onedot{.\@\xspace}
\def\eg{\emph{e.g}\onedot} 
\def\ie{\emph{i.e}\onedot}

\def\etal{\emph{et al}\onedot}

\newcommand{\be}{\begin{eqnarray}}
\newcommand{\ee}{\end{eqnarray}}
\newcommand{\bee}{\begin{eqnarray*}}
\newcommand{\eee}{\end{eqnarray*}}

\newcommand{\matrixb}{\left[ \begin{array}}
\newcommand{\matrixe}{\end{array} \right]}

\definecolor{green(ncs)}{rgb}{0.0, 0.62, 0.42}

\usepackage{amssymb}
\usepackage{pifont}
\newcommand{\cmark}{\ding{51}}%
\newcommand{\xmark}{\ding{55}}%
\usepackage{makecell}

\usepackage{xcolor}

\begin{document}
\twocolumn[{%
\renewcommand\twocolumn[1][]{#1}%
\maketitle 
\vspace{-3mm}
\begin{center}
\vspace{-5mm}
  \centering
\captionsetup{type=figure}
  \includegraphics[width=\linewidth]{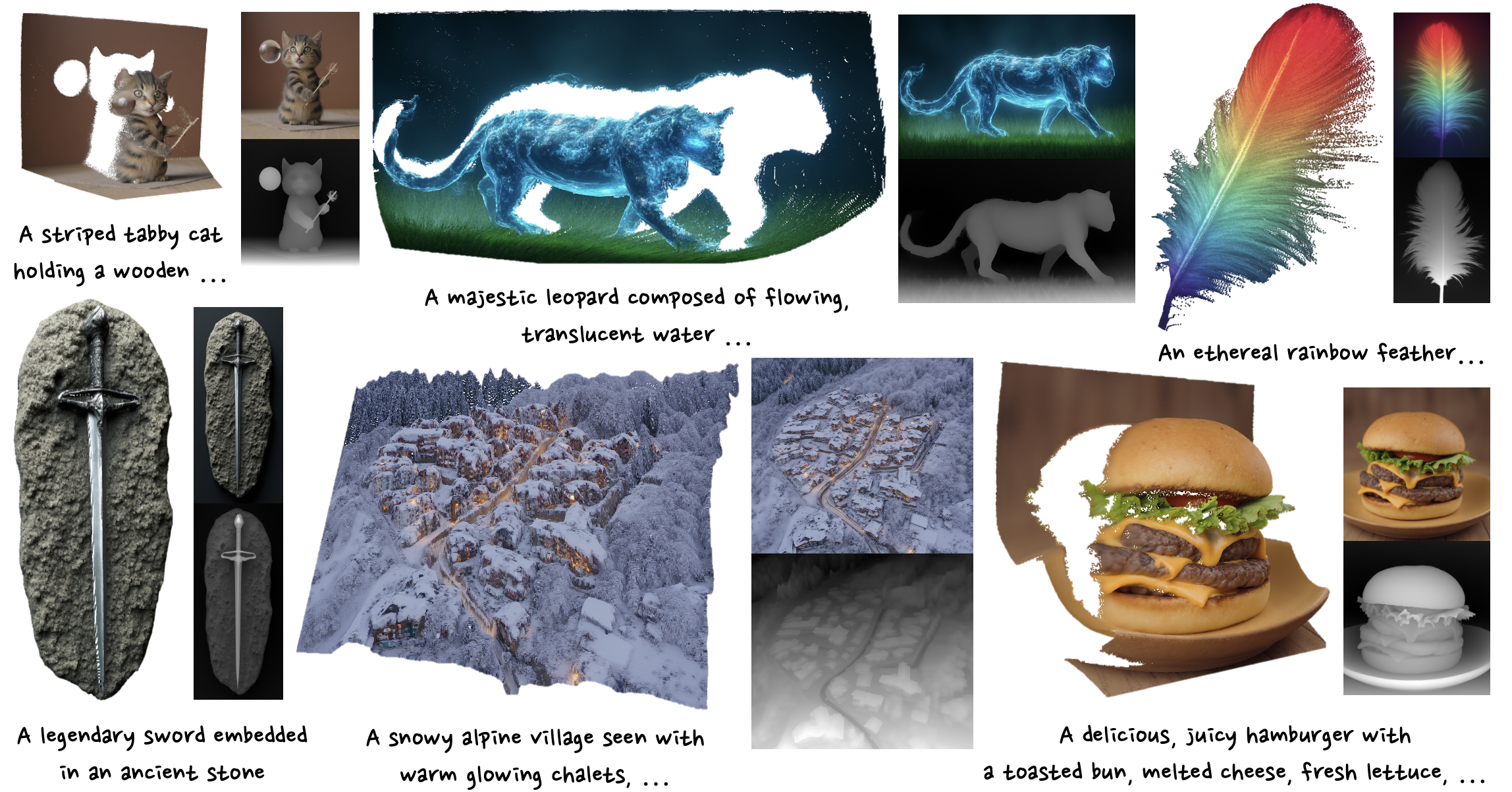}
  \vspace{-7mm}
  \captionof{figure}{
  We present JointDiT, a diffusion transformer modeling the RGB-Depth joint distribution.
  By leveraging the image prior of a state-of-the-art diffusion transformer~\cite{Flux2024}, JointDiT generates high-fidelity images and geometrically plausible and accurate depth maps.
  }
    \vspace{2mm}
    \label{fig:teaser}
\end{center}

}]
\def\thefootnote{$\dagger$}\footnotetext{Work done during an internship at Microsoft Research Asia.}
\begin{abstract}
We present JointDiT, a diffusion transformer that models the joint distribution of RGB and depth. 
By leveraging the architectural benefit and outstanding image prior of the state-of-the-art diffusion transformer, JointDiT not only generates high-fidelity images but also produces geometrically plausible and accurate
depth maps.
This solid joint distribution modeling is achieved through two simple yet effective techniques that we propose, namely, adaptive scheduling weights, which depend on the noise levels of each modality, and the unbalanced timestep sampling strategy.
With these techniques, we train our model across all noise levels for each modality, enabling JointDiT to naturally handle various combinatorial generation tasks, including joint generation, depth estimation, and depth-conditioned image generation by simply controlling the timesteps of each branch.
JointDiT demonstrates outstanding joint generation performance. Furthermore, it achieves comparable results in depth estimation and depth-conditioned image generation, suggesting that joint distribution modeling can serve as a viable alternative to conditional generation. The project page is available at \url{https://byungki-k.github.io/JointDiT/}
\end{abstract}    
\vspace{-4mm}
\begin{figure*}[t]
\centering
\includegraphics[width=1\linewidth]{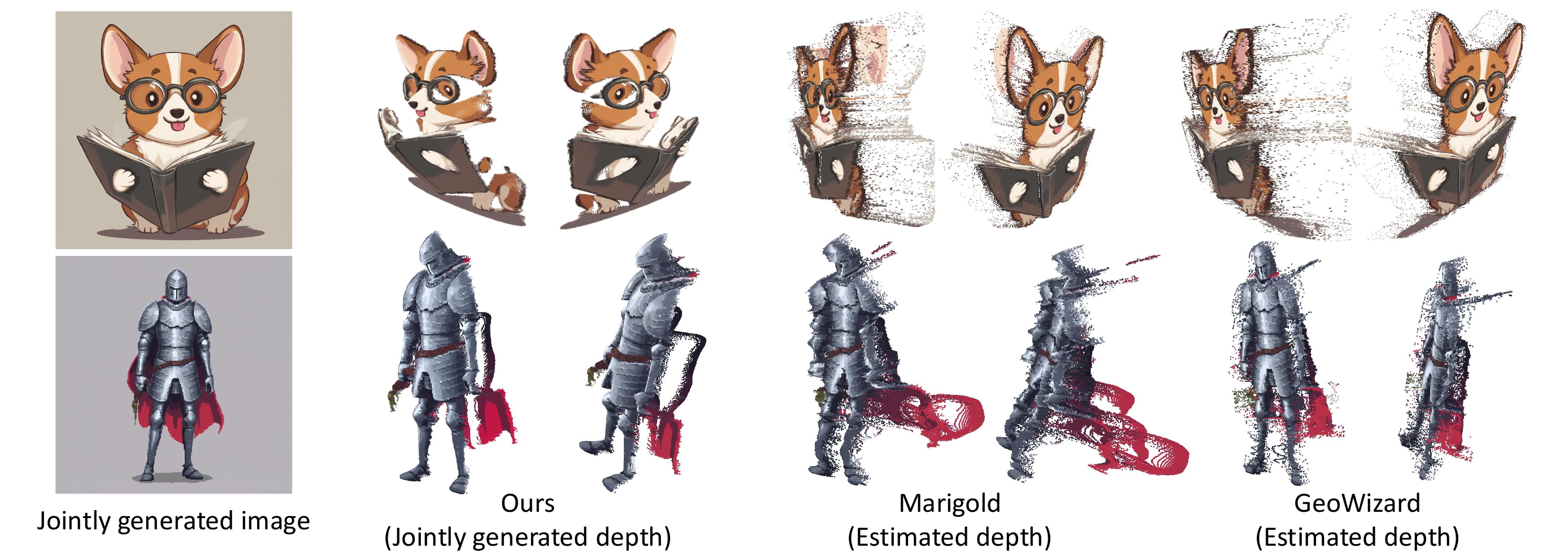}
    \caption{   
    \textbf{3D lifting results of JointDiT and depth diffusion models, \ie, Marigold~\cite{ke2024repurposing} and GeoWizard~\cite{fu2024geowizard}.} 
    JointDiT also shows more plausible 3D point clouds than depth estimation models in challenging illustration domains, likely due to the complementary behavior of the RGB and depth branches across generative processes, \ie, the RGB branch focuses on texture, and the depth branch on structure.
    } 
    \label{figure:depth_diffusion_3d}
\end{figure*}
\section{Introduction}
\label{sec:intro}

In the era of generative AI, diffusion models have made remarkable advancements in synthesizing images~\cite{ho2020denoising,rombach2022high}. 
The outstanding capability of text-to-image diffusion models has been found to be useful not only for image generation but also for solving important inverse problems~\cite{chung2022improving,chung2023solving,chung2024deep,hyoseok2025zero}, image inpainting~\cite{corneanu2024latentpaint,lugmayr2022repaint}, image editing~\cite{kawar2023imagic,couairon2022diffedit}, and even further cross-modal conditional generation, such as depth-conditioned image generation~\cite{zhang2023adding,bhat2024loosecontrol,luo2024readout}, and image-conditioned depth estimation~\cite{ke2024repurposing,fu2024geowizard,gui2024depthfm}.
These works have shown that using the image prior of diffusion models is effective for modeling conditional distribution.

Recently, in image and depth modalities, joint distribution modeling~\cite{stan2023ldm3d,zhang2023jointnet,li2024simple} has shown that it
not only enables joint generation but also shows potential as a viable alternative to existing depth estimation methods and depth-conditional image generation methods within a single unified framework by treating them as special cases of joint distribution modeling.
It demonstrates that joint modeling can be easily generalized for various tasks including controllable and conditional generation and estimation.
Despite this versatility, the realism of the generation is limited. 

In this work, we propose JointDiT, a diffusion transformer designed for solid joint distribution modeling of image and depth. Figure~\ref{fig:teaser} demonstrates the high-fidelity joint generation results of JointDiT. The high-fidelity images and geometrically accurate depth maps visually highlight the joint distribution capability of JointDiT, which has not been achieved.
Furthermore, we design JointDiT to provide a replaceable alternative to conditional distribution models by constructing a joint distribution at all noise levels for each modality. 
For instance, the model performs joint generation when both the image and depth map are noise, depth estimation when only the image is clean, and depth-conditioned image generation when only the depth map is clean.

To achieve this, we model the joint distribution by harnessing the strong image prior of a state-of-the-art diffusion transformer~\cite{Flux2024} and building a parallel depth branch 
 through joint connection modules.
By training on separate noise levels for each modality, JointDiT flexibly facilitates combinatorial tasks of image and depth by simply controlling the timestep of each branch.
To enable separate noise level training,
we propose two simple yet effective techniques, \ie, \textit{adaptive scheduling weights} and
\textit{unbalanced timestep sampling strategy}, 
designed for multi-modal diffusion training with separate noise levels. 
JointDiT achieves significantly superior joint generation results compared to previous joint generation methods~\cite{stan2023ldm3d,zhang2023jointnet,li2024simple} while demonstrating comparable performance in conditional generation tasks, such as depth estimation and depth-conditioned image generation. 
JointDiT also enables plausible depth generation even for challenging domains such as cartoon images and pixel art illustrations, where depth estimation methods~\cite{fu2024geowizard,ke2024repurposing} often struggle, as shown in Fig.~\ref{figure:depth_diffusion_3d}.
We further observe that the RGB and depth branches adopt complementary behavior in the joint generation process, with the depth branch capturing structural information and the RGB branch focusing on complementary
aspects related to texture and appearance, which may underlie the plausible results observed in challenging domains.

\vspace{2mm}
\noindent We summarize our contributions as follows:
\begin{itemize}
    \setlength{\itemsep}{1pt}
    \item We present JointDiT, a model for solid joint distribution modeling between image and depth modalities across all noise levels by leveraging the image prior of diffusion transformers. It supports combinatorial tasks, such as joint generation, depth estimation, and depth-conditioned image generation via simple timestep control.
    
    \item We propose adaptive scheduling weights and unbalanced timestep sampling strategy for separate noise level training in multi-modality, which significantly improves performance on combinatorial tasks.
    Through these techniques, we demonstrate that joint distribution modeling is a viable alternative to conditional generation.
\end{itemize}
\begin{figure*}[t]
\centering
\includegraphics[width=1\linewidth]{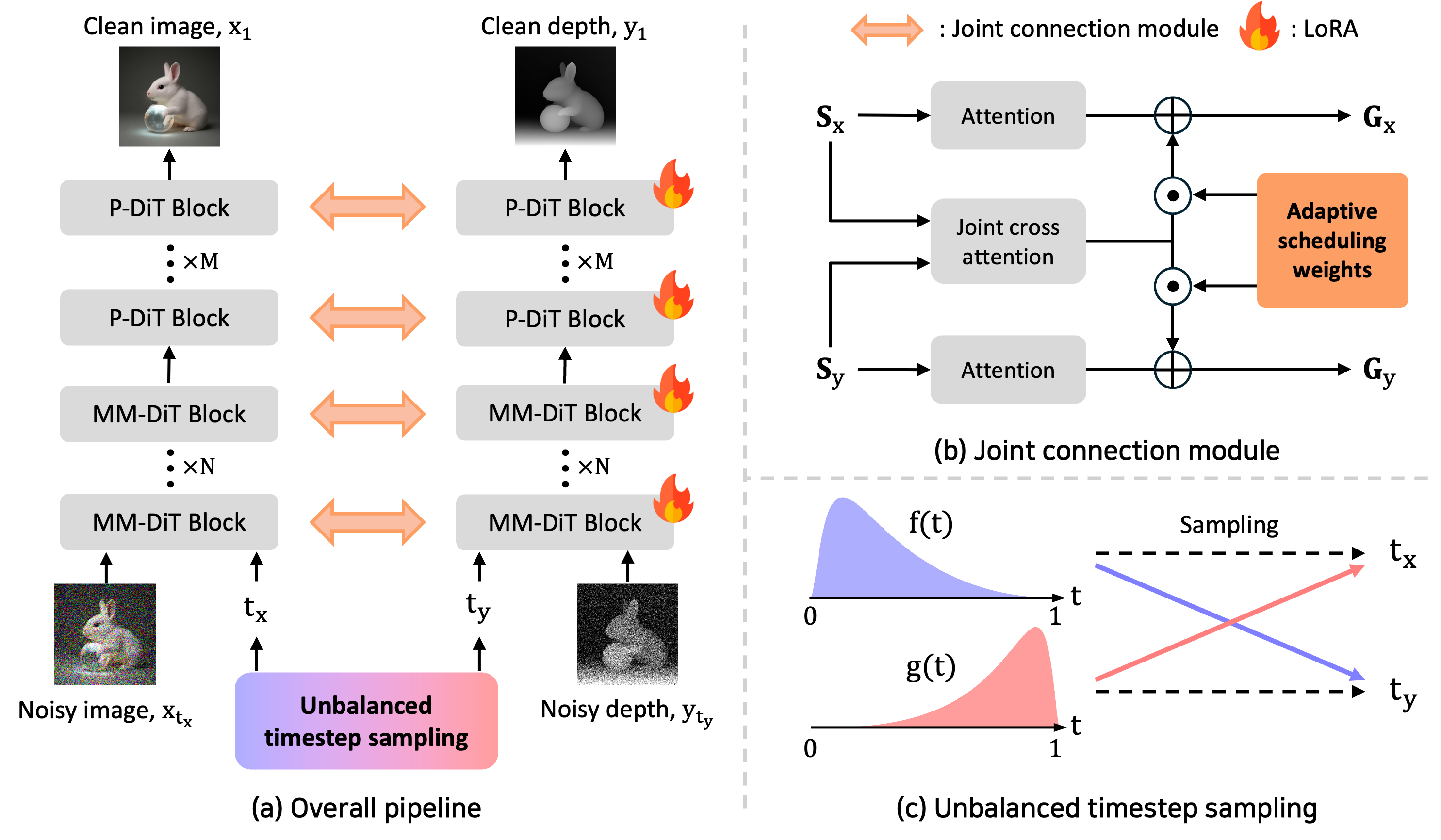}
    \caption{
    \textbf{Overall pipeline of JointDiT.} Building on Flux~\cite{Flux2024}, we introduce a parallel depth branch with trainable LoRAs~\cite{hu2022lora}. 
    The joint connection module enables the aligned joint generation. We propose adaptive scheduling weights and an unbalanced timestep sampling strategy for effective training using separate timesteps. MM-DiT and P-DiT denote the multi-modal~\cite{esser2403scaling} and parallel~\cite{dehghani2023scaling} diffusion transformers, respectively. The two timestep distributions, $f(t)$ and $g(t)$, are described in the supplementary material.
    }
    \label{figure:pipeline}
\end{figure*}

\section{Preliminaries}

\paragraph{Flow matching} 
Flow matching is generative modeling that learns a time-dependent vector field, which transports one probability distribution to another. It is closely related to Continuous Normalizing Flows (CNFs)~\cite{chen2018neural}, which model such transformations via differential equations.
We adopt the notation from Lipman~\etal~\cite{lipman2022flow} to describe the flow matching formulation and objective. Given the data points $x \in \mathbb{R}^d$ and probability density path $p: [0,1] \times \mathbb{R}^d \rightarrow \mathbb{R}_{>0}$ that is a time-dependent probability density function satisfying $\int p_{t}(x) dx = 1$, the time-dependent vector field $v: [0,1] \times \mathbb{R}^d \rightarrow \mathbb{R}^d$ combines with a flow $\phi: [0,1] \times \mathbb{R}^d \rightarrow \mathbb{R}^d$, leading to the ordinary differential equation (ODE):
\begin{equation}
    \frac{d}{d t}\phi_t(x) = v_t(\phi_t(x)), 
    \quad \phi_0(x) = x,
\end{equation}
where $\phi_0(x) = x$ is an initial condition. Through the push forward equation, the probability density function at time $t$, \ie, $p_t$, is transformed by $p_t = [\phi_t]_{*}p_0$. 
The $*$ and $p_0$ represent the push forward operator and simple prior, \eg, the standard normal distribution, and $p_{1}$ represents a data distribution. 
The objective of flow matching is to estimate $v_t(x)$ using a learnable neural network $v_{t, \theta}(x)$ by minimizing:
\begin{equation}
    \mathcal{L}_{\mathrm{FM}}
    =
    \mathbb{E}_{t,\, p_t(x)}
    \bigl[
        \|v_{t,\theta}(x) - v_t(x)\|
    \bigr].
\end{equation}
However, obtaining the true vector field $v_t$ is intractable. To address this, Lipman~\etal~\cite{lipman2022flow} proposed Conditional Flow Matching (CFM), which introduces a condition by sampling the accessible data sample $x_{1}$ from the unknown data distribution $q(x_1)$. 
By conditioning the true vector field on $x_{1}$, that is $v_{t}(x|x_{1})$, a tractable objective is obtained, as follows:
\begin{equation}
    \mathcal{L}_{\mathrm{CFM}}
    =
    \mathbb{E}_{t,\, q(x_1),p_t(x|x_1)}
    \bigl[
        \|v_{t,\theta}(x) - v_t(x|x_{1})\|
    \bigr].
\end{equation}

\section{Method}

Our goal is to develop a unified network that models the joint distributions between images and depth maps across all noise levels. 
This network can be applied to various tasks, including joint image-depth generation, depth estimation from an image, and depth-conditioned image generation by adjusting the noise levels of each modality.

To achieve this, inspired by previous works~\cite{li2024simple,chen2025diffusion} that employ separate noise sampling for images and conditions, we extend the flow matching framework to learn a joint vector field \( v_{t_x,t_y}(x,y|x_1,y_1) \) with two independent timesteps, \( t_x \) and \( t_y \). Here, \( x \) and \( y \) represent data points sampled from the RGB image and depth map distributions, respectively.
To estimate \( v_{t_x,t_y}(x,y|x_1,y_1) \), we design a learnable neural network \( v_{t_x,t_y,\theta}(x,y) \) and train it by minimizing the following Joint Conditional Flow Matching (JCFM) loss:
\begin{align}
    \mathcal{L}_{\mathrm{JCFM}}(\theta) &=
    \mathbb{E}_{t_x, t_y,q(x_1,y_1), p_{t_x,t_y}(x,y|x_1,y_1)} \Bigl[
    \|\,v_{t_x,t_y,\theta}(x,y) \nonumber \\
    &\quad -\; v_{t_x,t_y}(x,y|x_{1},y_{1})\|
    \Bigr].
\end{align}

Once the network successfully learns to estimate the vector field \( v_{t_x,t_y}(x,y|x_1,y_1) \), various tasks can be performed simply by adjusting $t_x$ and $t_y$ without any additional guidance. For example, initially setting \(t_x = 0, t_y = 0\) leads to the joint generation of both images and depth maps. When \(t_x=1, t_y=0\), it performs depth estimation from a given image, and when \(t_x = 0, t_y = 1\), it becomes a depth-conditioned image generation. In the later section, we will denote the noisy image and depth map samples $(x,y)\sim p_{t}(x,y)$ as $x_{t}$ and $y_{t}$ for simplicity.

\subsection{Joint Diffusion Transformer (JointDiT)}
Figure~\ref{figure:pipeline} shows JointDiT architecture. JointDiT is built on Flux~\cite{Flux2024}, an advanced diffusion transformer model that consists of multi-modal diffusion transformer (MM-DiT) and parallel diffusion transformer (P-DiT) blocks~\cite{esser2403scaling,dehghani2023scaling}. 
To harness its strong image prior and the benefits of transformer architectures in dense prediction tasks~\cite{ranftl2021vision}, we extend it to joint image and depth distribution modeling by introducing a parallel depth branch alongside the pre-trained RGB branch. 
Thereafter, we add LoRAs~\cite{hu2022lora} to the MM-DiT and P-DiT blocks to process the additional depth domain.
Additionally, joint connection modules are introduced in each DiT block to model joint distribution by interchanging features between the RGB and depth branches. We train the LoRAs and joint connection modules while keeping the pre-trained backbone model frozen.

In the joint connection modules (see Fig.~\ref{figure:pipeline}-b), feature exchange for joint distribution modeling occurs within the attention mechanism of each DiT block. 
We adopt the joint cross-attention module from the prior work~\cite{li2024simple}. 
This module facilitates joint distribution training by exchanging queries between the RGB and depth branches through attention mechanisms.
The motivation is that self-attention plays a key role in the form and structure of images~\cite{tumanyan2023plug}. 

To further reinforce this motivation, we propose adaptive scheduling weights, which encourage the joint model to follow the form and structure of the relatively cleaner domain between the RGB and depth branches. 
Specifically, we adaptively schedule the amount of information transferred between branches by joint cross attention according to the relative cleanliness of the given noisy image $x_{t_x}$ and the noisy depth $y_{t_y}$. 
This approach is also intuitively reasonable, as cleaner data inherently provides more useful information for joint generation.
The adaptive scheduling weights are individually multiplied with the joint cross-attention outputs, which corresponds to:
\begin{figure*}[t]
\centering
\includegraphics[width=1.0\linewidth]{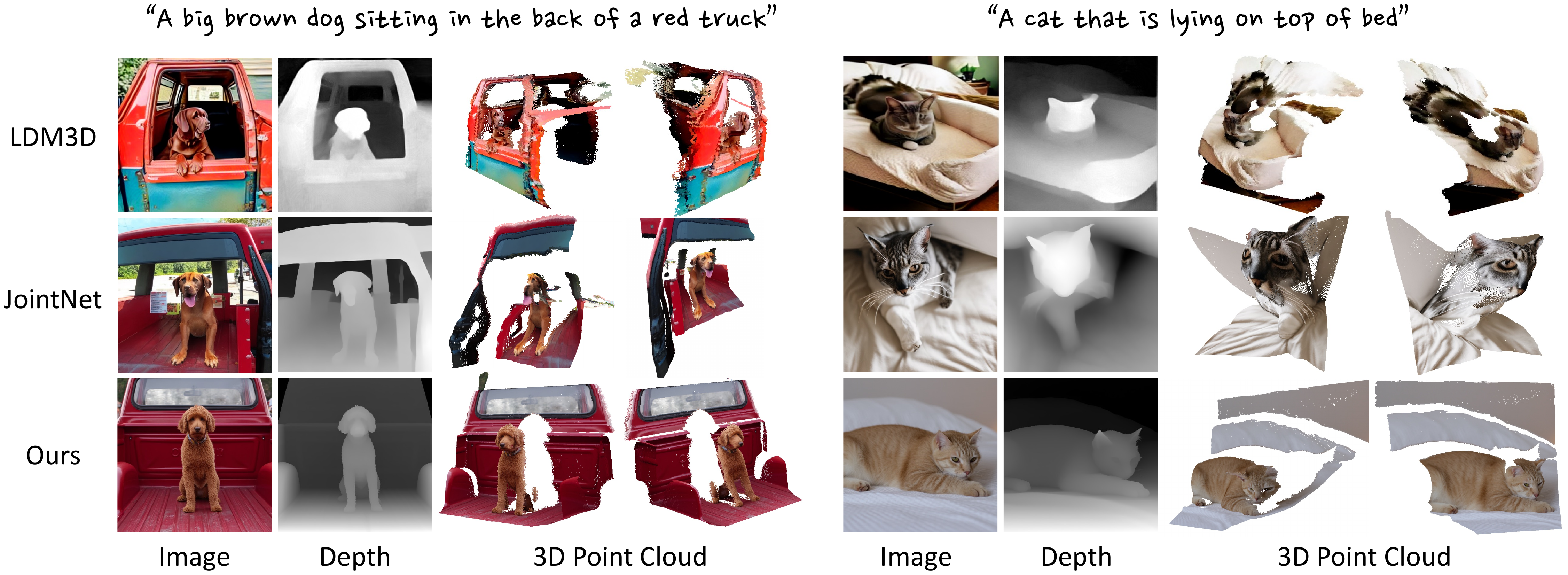}
    \caption{
    \textbf{3D lifting results of LDM3D~\cite{stan2023ldm3d}, JointNet~\cite{zhang2023jointnet}, and our JointDiT.}
    Our JointDiT generates highly plausible image-aligned 3D structures, surpassing previous joint generation methods in achieving superior consistency with real 3D space.
    } 
    \label{figure:joint_modeling_3d}
\end{figure*}
\begin{equation}
    \begin{aligned}
        \mathbf{G}_x &= \text{Attn}(\mathbf{S}_x) + w_x(t_x,t_y) \cdot \text{JointAttn}(\mathbf{S}_x,\mathbf{S}_y), \\
        \mathbf{G}_y &= \text{Attn}(\mathbf{S}_y) + w_y(t_x,t_y) \cdot \text{JointAttn}(\mathbf{S}_x,\mathbf{S}_y).
    \end{aligned}
\end{equation}
$w_x$ and $w_y$ are adaptive scheduling weights for:
\begin{equation}
    \begin{aligned}
        w_x(t_x,t_y) &= \text{sigmoid} \left( \alpha \left( \frac{t_y}{t_x + t_y} - \frac{1}{2} \right) \right), \\
        w_y(t_x,t_y) &= \text{sigmoid} \left( \alpha \left( \frac{t_x}{t_x + t_y} - \frac{1}{2} \right) \right),
    \end{aligned}
\end{equation}
where $\alpha$ is a scale factor. We set $\alpha$ to 3 for all experiments. 
The above equations indicate that more weight is given to the output of joint cross-attention for the noisier branch (relatively closer to $t=0$), letting it follow the domain structure of the cleaner branch.

We also introduce the unbalanced timestep sampling strategy to enforce joint distribution modeling at any separate timesteps (See Fig.~\ref{figure:pipeline}-c). Prior studies~\cite{esser2403scaling,zheng2024non} have investigated the impact of timestep sampling strategies on diffusion performance during training and have proposed various timestep sampling methods beyond uniform distribution.
Similarly, our base training code also employs a weighted timestep distribution for training\footnotemark[2] ($f(t)$ in Fig~\ref{figure:pipeline}-c).
\footnotetext[2]{https://github.com/kohya-ss/sd-scripts/tree/sd3}
However, with this timestep distribution, the joint distribution of $t_x$ and $t_y$ is likely insufficient to fully cover both joint generation and conditional generation tasks, as shown by Hang~\etal~\cite{hang2023efficient}, where insufficient timestep sampling negatively affected diffusion performance. The unbalanced timestep sampling strategy samples $t_{x}$ and $t_{y}$ independently from two unbalanced timestep distributions, $f(t)$ and $g(t)$, with half probability during training. For the remaining half, the same timesteps sampled from $f(t)$ are assigned to $t_x$ and $t_y$.
We experimentally validate that these two simple techniques are effective for building a solid joint distribution across all noise levels of images and depth maps, enhancing performance in joint generation, depth-conditioned image generation, and depth estimation.

\section{Experiments}
\label{sec:experiments}
We evaluate the performance of JointDiT across joint generation, depth estimation, and depth-conditioned image generation. We also analyze the behavior of the RGB and depth branches, as well as the effectiveness of adaptive scheduling weights and unbalanced timestep sampling. Experimental details can be found in the supplementary material.

\paragraph{Implementation details} To collect the training dataset, we randomly sample frames from a real-world internal video dataset, which allows us to acquire real-world images with a larger field of view easily. 
The sampled frames are resized while maintaining their aspect ratio, then center-cropped, to produce $512 \times 512$ images. 
The depth maps and text prompts are generated by Depth-Anything-v2~\cite{yang2025depth} and LLaVa~\cite{liu2023visual}, respectively. 
We train our model on the collected dataset, which consists of 50k pairs, for 75k iterations with a batch size of 4 and a learning rate of 1e-5. We use the LoRA rank of 64 in DiT blocks and apply text drop with a probability of 10\%~\cite{ho2022classifier}. The training is conducted on a single NVIDIA H100 GPU for 3.5 days.

\subsection{Joint Generation} 
We demonstrate JointDiT’s joint generation capability through visualizations of generated images, depth maps, and their 3D lifting results, which provide intuitive evidence of joint RGB–Depth modeling and underscore the necessity of joint distribution modeling.
To obtain the 3D lifting results, we apply an inverse projection to the generated image using the generated depth map. We compare our method with LDM3D~\cite{stan2023ldm3d} and JointNet~\cite{zhang2023jointnet}, as they provide raw depth maps that facilitate 3D lifting visualization. 

Figure~\ref{figure:joint_modeling_3d} demonstrates the results. Compared to LDM3D and JointNet, our JointDiT shows high-fidelity images, fine-detailed depth maps, and geometrically accurate 3D lifting results. In contrast, the 3D lifting results of JointNet and LDM3D are geometrically inaccurate. We assume that this significant gap in geometric accuracy is caused by the differences in the image prior and the architecture between the baseline models, \ie, stable diffusion~\cite{rombach2022high} and Flux~\cite{Flux2024}. The Flux model, which is built on the diffusion transformer architecture, demonstrates superior image generation quality over stable diffusion that adopts the UNet architecture. 
In addition, the transformer architecture has been shown effective in depth estimation by several studies~\cite{ranftl2021vision,li2021revisiting,agarwal2022depthformer} since it has the global receptive field different from the fully-convolutional networks. 

\begin{table*}[t]
    \centering        \resizebox{0.95\linewidth}{!}{
    \begin{tabular}{lllccccccccc}
        \toprule
        \multirow{2}{*}{Type} & \multirow{2}{*}{Method} & \multicolumn{2}{c}{NYUv2~\cite{silberman2012indoor}} & \multicolumn{2}{c}{ScanNet~\cite{dai2017scannet}} & \multicolumn{2}{c}{KITTI~\cite{behley2019semantickitti}} & \multicolumn{2}{c}{DIODE~\cite{vasiljevic2019diode}} & \multicolumn{2}{c}{ETH3D~\cite{schops2017multi}} \\
        \cmidrule(lr){3-4}\cmidrule(lr){5-6} \cmidrule(lr){7-8} \cmidrule(lr){9-10}  \cmidrule(lr){11-12}
        & & AbsRel $\downarrow$ & $\delta 1$ $\uparrow$ & AbsRel $\downarrow$ & $\delta 1$ $\uparrow$  & AbsRel $\downarrow$ & $\delta 1$ $\uparrow$ & AbsRel $\downarrow$ & $\delta 1$ $\uparrow$ & AbsRel $\downarrow$ & $\delta 1$ $\uparrow$ \\
        \midrule
        \multirow{6}{*}{Discriminative} 
        & MiDaS~\cite{ranftl2020towards} & 11.1 & 88.5 & 12.1 & 84.6 & 23.6 & 63.0 & 33.2 & 71.5 & 18.4 & 75.2 \\
        \multirow{6}{*}{depth estimation}& DPT~\cite{ranftl2021vision} & 9.8 & 90.3 & 8.2 & 93.4 & 10.0 & 90.1 & 18.2 & 75.8 & 7.8 & 94.6 \\
        & Depth-Anything-V2~\cite{yang2025depth} & 4.4 & 97.9 & 4.1 & 97.9 & 7.5 & 94.8 & 6.5 & 95.4 & 13.2 & 86.2 \\
        & 4M-21~\cite{bachmann20244m} & 11.8 & 88.7 & 10.6 & 89.1 & 15.6 & 78.9 & 32.1 & 75.1 & 8.4 & 93.8 \\
        & MultiMAE~\cite{bachmann2022multimae} & 9.2 & 91.6 & 8.7 & 92.3 & 16.9 & 75.1 & 35.2 & 71.9 & 10.6 & 89.9 \\
        & Unified-IO~\cite{lu2022unified} & 6.8 & 95.9 & 7.5 & 95.0 & 28.1 & 52.0 & 36.4 & 70.0 & 13.9 & 83.9 \\
        & Unified-IO 2~\cite{lu2024unified} & 12.5 & 85.8 & 14.5 & 81.6 & 48.9 & 31.7 & 43.4 & 62.0 & 20.5 & 72.1 \\
        \midrule
        \multirow{1}{*}{Generative} 
        & Marigold~\cite{ke2024repurposing} & 5.5 & 96.4 & 6.4 & 95.1 & 9.9 & 91.6 & 30.8 & 77.3 & 6.5 & 96.0 \\
        \multirow{1}{*}{depth estimation} & GeoWizard~\cite{fu2024geowizard} & 5.2 & 96.6 & 6.1 & 95.3 & \textbf{9.7} & \textbf{92.1} & 29.7 & \textbf{79.2} & \textbf{6.4} & 96.1 \\
        \midrule
        \multirow{3}{*}{Generative}
        & JointNet~\cite{zhang2023jointnet} & 13.7 & 81.9 & 14.7 & 79.5 & 20.9 & 66.7 & 35.0 & 58.5 & 27.1 & 73.5 \\
        \multirow{3}{*}{joint generation}  & UniCon~\cite{li2024simple} & 7.9 & 93.9 & 9.2 & 91.9 & --- & --- & --- & --- & --- & --- \\
        \cmidrule(lr){2-12}
        & Ours & 5.7 & 96.9 & 6.6 & 95.7 & 10.3 & 88.8 & 27.3 & 71.0 & 16.5 & 96.3 \\
        & Ours+ft & \textbf{5.0} & \textbf{97.3} & \textbf{5.6} & \textbf{96.5} & 10.9 & 87.7 & \textbf{26.6} & 71.1 & 9.3 & \textbf{96.8} \\
        \bottomrule
    \end{tabular}}
    \caption{\textbf{Depth estimation results.} We compare ours with generative joint generation methods, as well as with discriminative and generative depth estimation methods. Ours outperforms JointNet and UniCon. Additionally, it achieves comparable performance to generative depth estimation methods, except on ETH3D. \textbf{Bold} indicates the best performance among the generative methods in this table.
    }
    \label{tab:depth_results}
\end{table*}

\subsection{Depth Estimation}
\begin{figure}[t]
\centering
\includegraphics[width=1.0\linewidth]{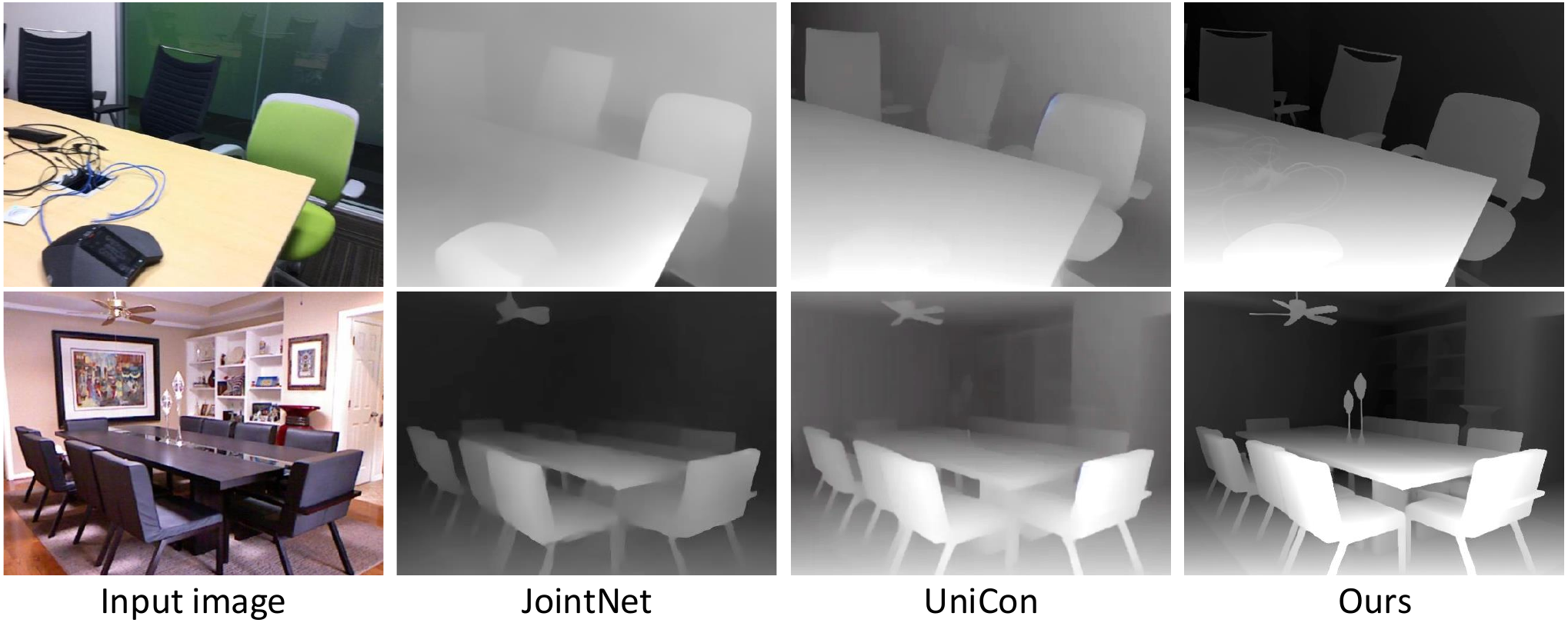}
    \caption{
    \textbf{Depth estimation results of joint generation models.} We visualize the depth estimation results of JointNet, UniCon, and our method on the NYUv2 and ScanNet dataset. 
    Our approach captures thin and fine-grained details with only a timestep adjustment, \ie, $t_{x}=1$ and $t_{y}=0$. 
    In contrast, JointNet requires additional fine-tuning for depth estimation.
    } 
    \label{figure:depth_estimation}
\end{figure}

We assess the depth estimation capability of JointDiT with different time steps, $t_{x}=1$ and $t_{y}=0$.
We compare our method with joint generation methods that support depth estimation, \eg, JointNet~\cite{zhang2023jointnet} and UniCon
~\cite{li2024simple}. 
We also compare with discriminative depth estimation methods, including depth-specialized models~\cite{ranftl2020towards,ranftl2021vision,yang2025depth} and multitask learning-based methods~\cite{bachmann20244m,bachmann2022multimae,lu2022unified,lu2024unified}, as well as generative depth estimation methods that utilize a diffusion model~\cite{ke2024repurposing,fu2024geowizard}.
Following the evaluation convention of prior work,
we compare each method on the NYUv2~\cite{silberman2012indoor}, ScanNet~\cite{dai2017scannet}, KITTI~\cite{behley2019semantickitti}, DIODE~\cite{vasiljevic2019diode}, and ETH3D~\cite{schops2017multi} datasets. The evaluation metrics are Absolute Mean Relative Error (AbsRel) and $\delta_1$.
Table~\ref{tab:depth_results} summarizes the results. Compared to joint generation methods, our model achieves superior performance across all evaluation datasets. Figure~\ref{figure:depth_estimation} shows the depth estimation results of joint generative methods on the ScanNet dataset. Compared to JointNet and UniCon, our method captures sharp edges and fine details.

We also compare our method with generative depth estimation models, which finetune most of the parameters of a pre-trained diffusion model. Except for the ETH3D dataset, our model achieves comparable performance with only a small portion of parameter tuning, \eg, LoRA layers and the joint connection module. On the ETH3D dataset, our method appears to achieve higher AbsRel than generative depth estimation methods, likely because we use the depth predictions of Depth-Anything-V2 for training.

To verify the depth estimation performance itself, we further trained our model for an additional 50k iterations on synthetic datasets. We collect the synthetic training dataset by filtering 80k data samples from Hypersim~\cite{roberts2021hypersim}, Replica~\cite{karaev2023dynamicstereo}, IRS~\cite{wang2021irs}, and MatrixCity~\cite{li2023matrixcity}. The training method remains the same as before. 
As shown in Tab~\ref{tab:depth_results}, the fine-tuned model, denoted as Ours+ft, achieved higher accuracy on the NYUv2, ScanNet, and DIODE datasets compared to generative depth estimation models. 
These results suggest two aspects.
First, the strong image prior and architectural properties of the diffusion transformer are effective for dense prediction tasks, even with only a small subset of trainable parameters.
Second, solid joint distribution modeling can serve as an alternative to conditional generation.

\begin{table}[t]
    \centering
    \begin{tabular}{lcc}
        \toprule
        \multirow{2}{*}{Method} & 
        \multicolumn{2}{c}{OpenImages 6K} \\
        \cmidrule(lr){2-3}
        & FID$\downarrow$ & AbsRel $\downarrow$ \\
        \midrule
        Readout-Guidance~\cite{luo2024readout} & 18.72 & 23.19 \\
        ControlNet~\cite{zhang2023adding}       & 13.68 & 9.85  \\
        UniCon~\cite{li2024simple}           & 13.21 & 9.26 \\
        \midrule
        Ours & \textbf{12.62} & \textbf{6.99} \\
        \bottomrule
    \end{tabular}
    \caption{\textbf{Depth-conditioned image generation results.} With the same training dataset, ours achieves the lowest FID and AbsRel.}
    \label{tab:depth_conditioned_image}
\end{table}
\begin{table*}[h]
    \centering
    \resizebox{0.93\linewidth}{!}{
    \begin{tabular}{ccccccccccc}
        \toprule
        \multirow{2}{*}{Adaptive} & \multirow{2}{*}{Unbalanced} & \multicolumn{3}{c}{ImageNet 6K} & \multicolumn{3}{c}{Pexels 6K} & \multicolumn{3}{c}{MSCOCO 30K} \\
        \cmidrule(lr){3-5} \cmidrule(lr){6-8} \cmidrule(lr){9-11}
        \multirow{-1}{*}{scheduling weights} & \multirow{-1}{*}{timestep} & FID$\downarrow$ & IS$\uparrow$ & CLIP$\uparrow$ & FID$\downarrow$ & IS$\uparrow$ & CLIP$\uparrow$ & FID$\downarrow$ & IS$\uparrow$ & CLIP$\uparrow$ \\
        \midrule
        \textcolor{red}{\xmark} & \textcolor{red}{\xmark} & 30.88 & 31.61 & 29.80 & 21.85 & 20.02 & 30.53 & 15.17 & 29.73 & 30.21 \\
        \textcolor{red}{\xmark} & \textcolor{green}{\cmark} & 29.37 & 33.36 & 29.89 & 22.01 & 19.68 & 30.15 & 13.76 & 30.73 & 30.43 \\
        \textcolor{green}{\cmark} & \textcolor{red}{\xmark} & \textbf{24.20}	& 37.04 & 30.37 & \textbf{19.49} & 21.60 & 30.53 & \textbf{11.13} & 33.98 & 30.63 \\
        \textcolor{green}{\cmark} & \textcolor{green}{\cmark} & 24.26 & \textbf{37.81} & \textbf{30.51} & 19.87 & \textbf{22.51} & \textbf{30.71} & 11.27 & \textbf{34.35} & \textbf{30.76} \\
        \bottomrule
    \end{tabular}}
    \caption{\textbf{Ablation studies on joint generation.} Applying adaptive scheduling weights notably improves all evaluation metrics across all datasets. The unbalanced timestep sampling strategy enhances IS and CLIP scores when combined with adaptive scheduling weights.
    }        
    \label{tab:ablation_joint}
\end{table*}

\begin{table*}[ht]
    \centering
    \resizebox{0.93\linewidth}{!}{
    \begin{tabular}{cccccccccccc}
        \toprule
        \multirow{2}{*}{Adaptive} & \multirow{3}{*}{Unbalanced} & \multicolumn{2}{c}{NYUv2} & \multicolumn{2}{c}{ScanNet} & \multicolumn{5}{c}{OpenImages 6K} \\
        \cmidrule(lr){3-4} \cmidrule(lr){5-6} \cmidrule(lr){7-11}
        \multirow{1}{*}{scheduling} & \multirow{2}{*}{timestep} & \multirow{2}{*}{AbsRel $\downarrow$} & \multirow{2}{*}{$\delta_1$ $\uparrow$} & \multirow{2}{*}{AbsRel $\downarrow$} & \multirow{2}{*}{$\delta_1$ $\uparrow$} & \multirow{2}{*}{FID$\downarrow$} & \multicolumn{4}{c}{ImageReward} \\
        \cmidrule(lr){8-11}
        \multirow{-2}{*}{weights} & & & & & & & Rank1$\uparrow$ & Rank2 & Rank3 & Rank4$\downarrow$ \\
        \midrule
        \textcolor{red}{\xmark} & \textcolor{red}{\xmark} & 8.8 & 92.2 & 10.4 & 88.9 & \textbf{11.94} & 26.35 &  26.38 &	24.43 &	22.83 \\
        \textcolor{red}{\xmark} & \textcolor{green}{\cmark} & 7.8 & 93.9 & 8.7 & 92.4 & 12.51 & 21.77 &	24.82 &	25.48 &	27.93\\
        \textcolor{green}{\cmark} & \textcolor{red}{\xmark} & 6.4 & 96.0 & 7.2 & 94.9 &  14.37 & 21.15 &	22.73 &	25.63 &	30.48 \\
        \textcolor{green}{\cmark} & \textcolor{green}{\cmark} & \textbf{5.7} & \textbf{96.9} & \textbf{6.6} & \textbf{95.7} &  12.58 & \textbf{30.73} &	26.07 & 24.45 & \textbf{18.75} \\
        \bottomrule
    \end{tabular}}
    \caption{\textbf{Ablation studies on depth estimation and depth-conditioned image generation.} Adaptive scheduling weights and unbalanced timestep sampling are effective for depth estimation. In depth-conditioned image generation, using both methods together achieved the best performance in ImageReward~\cite{xu2023imagereward} ranking, trained to capture human preference, with the first rank highest and the last ranking lowest.
    }
    \label{tab:ablation_simple}
\end{table*}

\subsection{Depth-Conditioned Image Generation} 
We validate the depth-conditioned image generation quality, another joint generation with different time steps, $t_{x}=0$ and $t_{y}=1$.
Following the evaluation protocol of UniCon~\cite{li2024simple}, 
We compare our method with Readout-Guidance~\cite{luo2024readout}, ControlNet~\cite{zhang2023adding}, and UniCon.
All methods are trained on the same dataset, \ie, 16k samples from PascalVOC~\cite{everingham2010pascal}, and evaluated on 6k samples from OpenImages~\cite{krasin2017openimages}.
The evaluation is based on the FID score between the generated and original images, as well as the consistency of depth estimation results, \eg, AbsRel. 
Table~\ref{tab:depth_conditioned_image} shows the depth-conditioned image generation results on the OpenImages 6K dataset. Compared to other methods, our method shows a lower FID score and AbsRel. 
The lower AbsRel indicates that the generated images accurately preserve the original image’s geometry.

\begin{figure}[t]
\centering
\includegraphics[width=0.95\linewidth]{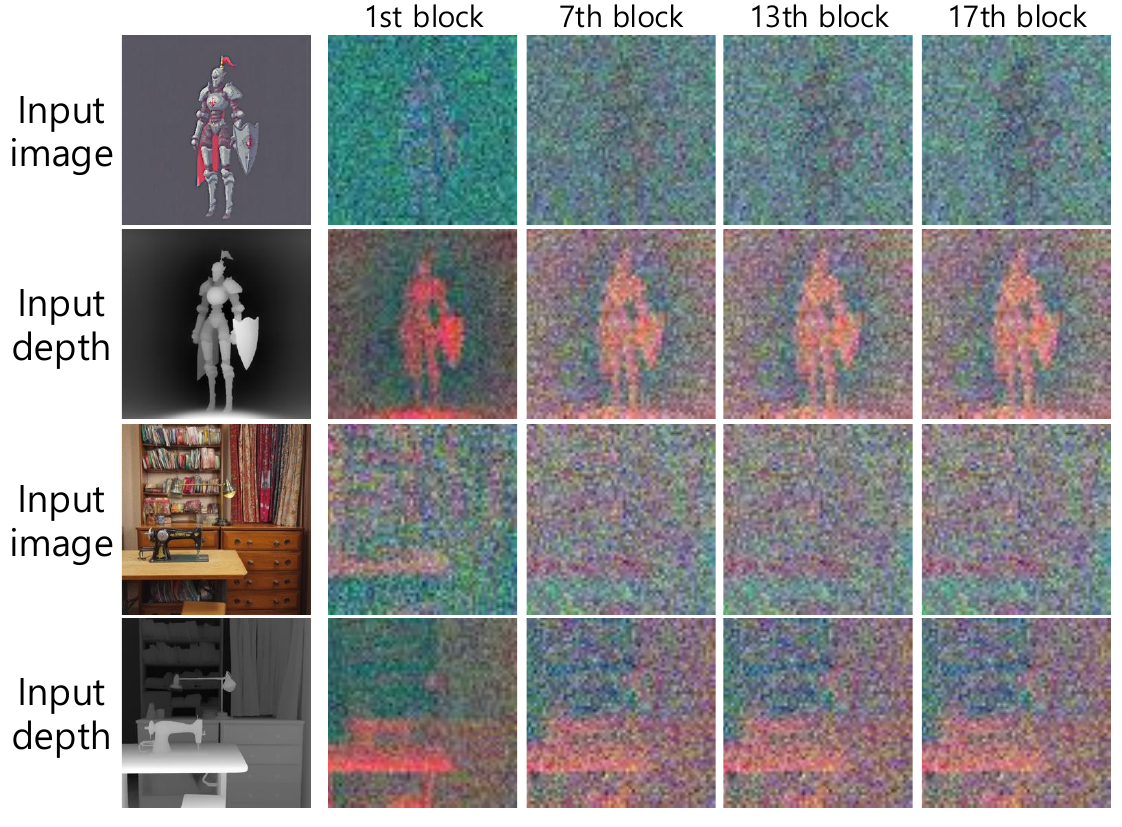}
    \vspace{-0.3cm}
    \caption{
    \textbf{Feature visualization of RGB-Depth branches in MM-DiT blocks.}
    We observe that, at timestep $t=0.48$, the depth branch tends to focus on scene geometry, while the RGB branch captures semantic patterns related to texture and appearance.
    } 
    \vspace{-0.3cm}
    \label{figure:features}
\end{figure}

\subsection{Ablation Studies}

\paragraph{Joint RGB-Depth feature visualization}
As shown in Fig.~\ref{figure:depth_diffusion_3d}, joint RGB-Depth modeling enables plausible depth generation from text prompts, even for stylized domains such as cartoon-style and pixel art illustrations. Furthermore, it tends to yield more structured and plausible 3D lifting results, not only for stylized domains but also for realistic image domains, compared to depth prediction from RGB images alone, as can be found in the supplementary material. 
The way the RGB and depth branches focus on textures and structure, respectively, may hint at how such capability achieves.
Figure~\ref{figure:features} shows the joint RGB-Depth features of MM-DiT blocks, visualized following the method of Tumanyan~\etal~\cite{tumanyan2023plug}. The depth branch focuses on capturing underlying structural information, while the RGB branch attends to complementary aspects related to texture and appearance. This behavior may suggest that the depth branch, by focusing on geometric properties throughout the joint generation process, enables plausible depth generation even in scenes where depth estimation is challenging.

\paragraph{Adaptive scheduling weights and unbalanced timestep sampling}
We analyze the effectiveness of adaptive scheduling weights and unbalanced timestep sampling in achieving solid joint distribution modeling.
For comparison, we train four models on our dataset for 75k iterations, varying the use of adaptive scheduling weights and unbalanced timestep sampling by either applying or omitting. When not using unbalanced timestep sampling, we respectively sample $t_{x}$ and $t_{y}$ from timestep distribution $f(t)$, which is the distribution that our base training code suggests.
We first investigate the effect in joint generation by assessing the quality of images generated from text prompts on three datasets: 30k samples from MS-COCO~\cite{lin2014microsoft}, 6k from ImageNet~\cite{russakovsky2015imagenet}, and 6k from Pexels~\cite{Pexels}.
As evaluation metrics, we use the Inception Score (IS)~\cite{salimans2016improved}, Fréchet Inception Distance (FID)~\cite{heusel2017gans}, and CLIP similarity~\cite{radford2021learning} as evaluation metrics.
Table~\ref{tab:ablation_joint} shows that the usage of adaptive scheduling weights significantly improves all metrics across all evaluation datasets. When unbalanced timestep sampling is applied together, IS and CLIP scores tend to improve. These results suggest that considering the relative noise level is important for effectively connecting different modality generation branches in separate timestep training.

We also evaluate each model on depth estimation and depth-conditioned image generation to assess the joint distribution modeling performance at the most extreme timesteps, specifically at $t_{x}=0$ and $t_{y}=1$, and vice versa. We use the NYUv2, ScanNet, and OpenImages 6K datasets to evaluate the performance of depth estimation and depth-conditioned image generation. 
Table~\ref{tab:ablation_simple} reports the results. In depth estimation, applying either adaptive scheduling weights or unbalanced timestep sampling improves performance. The best results are achieved when both are used together.
For the depth-conditioned image generation, we follow UniCon~\cite{li2024simple} for the evaluation setting, using 6K samples from the OpenImages dataset. 
We report the FID between original and depth-conditioned images, and also the percentage of samples ranked 1st to 4th by ImageReward~\cite{xu2023imagereward}, a human preference-trained reward model for text-to-image generation.
As shown in Tab~\ref{tab:ablation_simple}, applying only adaptive scheduling weights results in lower performance in terms of FID. However, when combined with unbalanced timestep sampling, the performance becomes comparable to other configurations.
Notably, when adaptive scheduling weights and unbalanced timestep sampling are applied together, ImageReward ranking 1 has the highest proportion, while the last ranking has the lowest. These results demonstrate that the two techniques effectively model the joint distribution at extreme timesteps.

\section{Related Work}
\paragraph{Text-to-image diffusion models} 
The success of DDPM~\cite{ho2020denoising} has demonstrated the effectiveness of diffusion models for text-to-image generation, utilizing a forward and backward process formulated as a Markov chain. Score-based generative models~\cite{song2019generative,song2020improved} provided another perspective by modeling the diffusion process as learning the score, \ie, the gradient of the log probability density, from noisy data. It was further extended with Stochastic Differential Equations (SDEs), which unify the forward and backward processes in a continuous-time framework~\cite{song2020score}. More recently, Flow Matching~\cite{lipman2022flow} has been introduced as an alternative to diffusion models, enabling exact likelihood training through Continuous Normalizing Flows (CNFs)~\cite{chen2018neural}.

Stable Diffusion~\cite{rombach2022high} improved the efficiency of the diffusion process by operating in a latent space instead of the image space, allowing for a more compact and expressive representation. This approach demonstrated impressive results. While early diffusion models primarily relied on U-Net architectures, recent studies have shown that the transformer-based architecture~\cite{vaswani2017attention} can also be highly effective for diffusion models. The diffusion transformer~\cite{peebles2023scalable} benefits from the global receptive field and scalability of the transformers, leading to improved generation quality. Models such as Flux and PixArt-$\alpha$~\cite{chen2023pixart} further demonstrate these advantages, highlighting the potential of transformers in text-to-image generation.

\paragraph{Joint and conditional diffusion models}
Text-to-image diffusion models have demonstrated their effectiveness in conditional and joint generation tasks. ControlNet~\cite{zhang2023adding} introduced an additional zero-initialized network, enabling fine-grained control over conditional generation tasks such as depth-to-image and pose-to-image synthesis. Building on this, LooseControl~\cite{bhat2024loosecontrol} proposed a more relaxed conditioning approach, allowing for weaker or more flexible integration of conditional information. 
Meanwhile, the image prior of pre-trained diffusion models can be beneficial in a wide range of computer vision tasks, such as sound-to-image generation~\cite{sung2025soundbrush,sung2024sound2vision}, single-image 3D reconstruction~\cite{liu2023zero,yu2024metta}, and 3D object texturing~\cite{chen2023text2tex,youwang2024paint}.

In depth-related tasks, prior works~\cite{ke2024repurposing,fu2024geowizard,gui2024depthfm} have demonstrated the effectiveness of diffusion priors for depth estimation.
Hyoseok~\etal~\cite{hyoseok2025zero} further showed that such priors can be used to solve inverse problems, specifically depth completion.
Models such as JointNet~\cite{zhang2023jointnet}, LDM3D~\cite{stan2023ldm3d}, and UniCon~\cite{li2024simple} were designed to model the joint distribution between images and depth maps using diffusion models. However, these models are based on a U-Net based diffusion architecture, which has a limited receptive field. This is in contrast to recent findings suggesting that diffusion transformers provide a stronger image prior and a global receptive field, which is particularly useful for dense prediction tasks~\cite{ranftl2021vision,li2021revisiting,agarwal2022depthformer}. In this paper, we leverage the advantages of diffusion transformer to model the solid joint distribution between image and depth.

\section{Conclusion}
We propose JointDiT, which models solid joint distribution. By harnessing the strong image prior and global receptive property of a state-of-the-art diffusion transformer, we build a unified model capturing multi-modal joint distribution at any separate noise levels. 
To achieve solid distribution, we present two simple yet effective techniques, called the adaptive scheduling weights dependent on the noise levels of modalities and unbalanced timestep sampling strategy. Through comprehensive experiments, we show that these two techniques notably improve the performance of joint generation, depth estimation, and depth-to-image generation. 
Our complete model generates images and depth maps, which form highly plausible and image-aligned 3D structures when lifted into 3D space. Furthermore, JointDiT achieves depth estimation performance comparable to that of diffusion-based depth estimation models, demonstrating that a joint distribution model can serve as a viable alternative for conditional distribution models.

\paragraph{Limitation}
To generate an image and its corresponding depth map, JointDiT requires an input batch size of 2 and additional parameters, resulting in a 19.8\% increase in network parameters and a 2.9$\times$ longer sampling time for 20 steps.
Exploring adaptations to a lightweight diffusion transformer model would be a promising research direction.

\section*{Acknowledgment}
This work was supported by the MSIT (Ministry of Science, ICT), Korea, under the Global Research Support Program in the Digital Field program (RS-2024-00436680) supervised by the IITP (Institute for Information \& Communications Technology Planning \& Evaluation). It was also supported by the KAIST cross-generation collaborative lab project, the Ministry of Science and ICT and NIPA (HPC Support Project), and Microsoft Research Asia.

{
    \small
    \bibliographystyle{ieeenat_fullname}
    \bibliography{main}
}

\appendix
\clearpage
\setcounter{page}{1}
\maketitlesupplementary

\hypersetup{linkcolor=black}

\hypersetup{linkcolor=black}
\section*{Contents}
\textbf{A \ \ \ Implementation Details}\\
\text{\qquad A.1 \ \ \ Experiment Setup} \\ 
\text{\qquad A.2 \ \ \ Data Preprocessing} \\ 
\text{\qquad A.3 \ \ \ Unbalanced Timestep Sampling Strategy} \\
\text{\qquad A.4 \ \ \ Architecture of JointDiT} \\
\textbf{B \ \ \ Additional Experiments} \\
\text{\qquad B.1 \ \ \ 
Advantages of Joint RGB-Depth Modeling
} \\ 
\text{\qquad B.2 \ \ \ 
Jointly Generated Image Quality
} \\ 
\text{\qquad B.3 \ \ \ 
Ablation of the LoRA's Rank
} \\ 
\text{\qquad B.4 \ \ \ 
Analysis of Failure Cases} \\ 
\text{\qquad B.5 \ \ \ 
Joint Panorama Generation
} \\ 
\textbf{C \ \ \ Additional Qualitative Results} \\ 
\noindent\rule{\linewidth}{0.2pt}
\hypersetup{linkcolor=blue}

\section{Implementation Details}
We provide the details of the experiment setup, dataset preprocessing, proposed unbalanced timestep sampling strategy, and architecture design of JointDiT.

\subsection{Experiment Setup}
We will describe in detail the configurations we used for joint generation, depth estimation, depth-conditioned image generation, and Joint RGB-Depth feature visualization. We consistently use 20 denoising steps across all experiments.

\paragraph{Joint generation}
We generate images and their corresponding depth maps by initially setting $t_{x}=0$ and $t_{y}=0$ by sampling noises from a standard normal distribution. 
While the main paper presents joint generation results conditioned on text prompts, we find that joint generation occurs even without a text prompt.
To compare with JointNet~\cite{zhang2023jointnet} and LDM3D~\cite{stan2023ldm3d}, we generate 512×512 images and depth maps jointly. Despite being trained only on a 512×512 resolution dataset, we observe that JointDiT successfully operates at varying resolutions, such as 1024×1024.

\begin{table}[t]
    \centering
    \begin{tabular}{cc}
        \toprule
        Type & LoRA applied components \\
        \midrule
        \multirow{6}{*}{MM-DiT} 
        & img\_mod.lin \\
        & img\_attn.qkv \\
        & txt\_mod.lin \\
        & txt\_attn.qkv \\
        & img\_attn.proj \\
        & txt\_attn.proj \\
        \midrule
        \multirow{2}{*}{P-DiT} 
        & linear1 \\
        & modulation.lin \\
        \midrule
        \multirow{3}{*}{Input stage} 
        & vector\_in.in\_layer \\
        & vector\_in.out\_layer \\
        & txt\_in \\
        \bottomrule
    \end{tabular}
    \caption{\textbf{LoRA-applied components.} To build the depth branch extending the original Flux model~\cite{Flux2024}, we add LoRAs to MM-DiT, P-DiT, and Input stage.}
    \label{tab:lora_flux}
\end{table}

\paragraph{Depth estimation}
To estimate the depth map from a given image, we set $t_{x}=1$ and $t_{y}=0$ and provide an empty text prompt. 
Unlike Marigold~\cite{ke2024repurposing} and Geowizard~\cite{fu2024geowizard}, we do not use any ensemble technique.
Since JointDiT can operate at varying resolutions, we use the NYUv2, ScanNet, KITTI, and DIODE datasets~\cite{silberman2012indoor,dai2017scannet,behley2019semantickitti,vasiljevic2019diode} at their original resolutions as model inputs. For the ETH3D dataset~\cite{schops2017multi}, which has a 4K resolution, we resize the images while preserving the aspect ratio so that the larger dimension is set to 1024 pixels. This preprocessing strategy is consistently applied to the comparison methods as well, and for methods that require a fixed input resolution, we use their designated resolution for evaluation.

\paragraph{Depth-conditioned image generation}
We generate depth-conditioned images from given text prompts by initially setting $t_{x}=0$ and $t_{y}=1$. The conditioning depth maps are obtained by Depth-Anything-V2~\cite{yang2025depth}. 
For the experiment of Sec.~4.3 in the main paper, we follow the evaluation setting of UniCon~\cite{li2024simple} to compare with Readout-Guidance~\cite{luo2024readout}, ControlNet~\cite{zhang2023adding}, and UniCon. Specifically, we train our model and these methods on the same training dataset, which includes 16k images of PascalVOC~\cite{everingham2010pascal}, depth maps from Depth-Anything-V2~\cite{yang2025depth}, and text prompts extracted using BLIP2~\cite{li2023blip}. 
For the evaluation, using the selected 6k images from the OpenImages dataset~\cite{krasin2017openimages}, we estimate depth maps using an off-the-shelf model and generate images conditioned on these depth maps and text prompts from BLIP2.

\paragraph{Joint RGB-Depth feature visualization}
For the feature visualization of Sec.~4.4 in the main paper, we strictly follow the method proposed by Tumanyan~\etal~\cite{tumanyan2023plug}, and visualize the PCA results of the features from each MM-DiT block.
Similar to Tumanyan~\etal, who collected images from semantically related domains (such as humanoid pictures) for visualization, we perform joint generation on 50 samples for each domain, \ie, pixel art style illustrations and indoor scenes that are used in the two examples shown in Fig.~6 of the main paper.
We extract features at approximately 50\% of the generation process (\ie, $t = 0.48$), and apply PCA to visualize them.
Due to the architecture structure of the Flux model, which applies positional encoding immediately before every attention layer, we subsample the even indices before applying PCA.

\subsection{Data Preprocessing}
We randomly sample RGB frames from the internal video dataset, which has a resolution of 512×512 or higher. The sampled frames are resized so that the smaller dimension (width or height) is 512 pixels, followed by a 512×512 center crop.
We obtain text prompts from the 512×512 images using LLaVA~\cite{liu2023visual}. To generate the corresponding disparity maps, we use Depth-Anything-V2 and normalize them so that the maximum and minimum values are 1 and 0, respectively.

\begin{figure*}[t]
\centering
\includegraphics[width=0.96\linewidth]{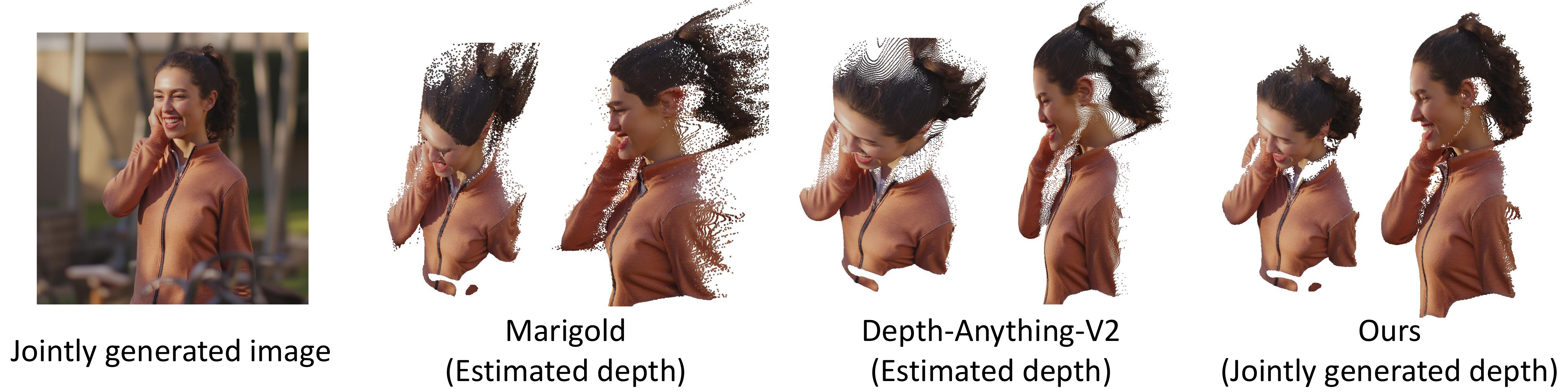}
    \caption{
    \textbf{Comparison of 3D lifting results from our JointDiT, Marigold, and Depth-Anything-V2.} The jointly generated depth from JointDiT leads to more coherent 3D shapes and better preservation of structural details compared to the estimated depths.
    } 
    \label{figure:realistic}
\end{figure*}
\begin{table*}[t]
    \centering
    \resizebox{0.98\linewidth}{!}{
    \begin{tabular}{llccccccccc}
        \toprule
        \multirow{1.5}{*}{Generation} & \multirow{2}{*}{Method} & \multicolumn{3}{c}{ImageNet 6K} & \multicolumn{3}{c}{Pexels 6K} & \multicolumn{3}{c}{MSCOCO 30K} \\
        \cmidrule(lr){3-5} \cmidrule(lr){6-8} \cmidrule(lr){9-11}
        \multirow{0.1}{*}{modality} & & FID$\downarrow$ & IS$\uparrow$ & CLIP$\uparrow$ & FID$\downarrow$ & IS$\uparrow$ & CLIP$\uparrow$ & FID$\downarrow$ & IS$\uparrow$ & CLIP$\uparrow$ \\
        \midrule
        \multirow{2}{*}{Image} & SD v2.1~\cite{rombach2022high} & 23.13 & 40.49 & 31.16 & 20.53 & 24.73 & 31.37 & 15.00 & 37.13 & 31.37 \\
        & Flux & 25.96 & 46.12 & 30.90 & 24.71 & 25.32 & 31.09 & 22.85 & 41.40 & 30.77 \\
        \midrule
        & JointNet~\cite{zhang2023jointnet} & 25.92 & 37.23 & 30.50 & 20.28 & 24.94 & 30.72 & 12.62 & 35.88 & 30.80 \\
        Image \& depth & LDM3D~\cite{stan2023ldm3d} & 37.72 & 31.73 & 30.45 & 32.50 & 20.26 & 30.52 & 25.58 & 29.36 & 30.81 \\
        \cmidrule(lr){2-11}
        & Ours & 24.26 & 37.81 & 30.51 & 19.87 & 22.51 & 30.71 & 11.27 & 34.35 & 30.76 \\
        \bottomrule
    \end{tabular}}
    \caption{\textbf{Quantitative evaluation on jointly generated images.} We present the performance of the baseline model for comparison. Our method achieves performance comparable to JointNet, while LDM3D demonstrates relatively poor results. Compared to our base model, \ie, Flux, we achieve lower FID scores but also lower IS scores, likely due to the limited size of the training dataset.
    }        
    \label{tab:quan_joint}
\end{table*}

\paragraph{Synthetic dataset} We further fine-tune our model to verify the depth estimation capability itself. We utilize the Hypersim~\cite{roberts2021hypersim}, Replica~\cite{karaev2023dynamicstereo}, IRS~\cite{wang2021irs}, and MatrixCity~\cite{li2023matrixcity} datasets for fine-tuning. We first unify the ground-truth depth or disparity maps of the synthetic datasets into disparity maps because our model was previously trained on the disparity maps of Depth-Anything-V2. Thereafter, we define invalid regions for each dataset. For example, in MatrixCity, the depth of the sky was set to the maximum value, while in Replica, there exist depth values that are closer than the camera plane. Then, we apply the bias and scale to the ground-truth disparity map so that the mean and standard deviation match those of Depth-Anything-V2's disparity estimation at valid regions. The annotations in invalid regions are replaced with Depth-Anything-V2's estimation. This process allows us to obtain annotations for invalid regions while ensuring consistency in depth map characteristics, which can vary significantly when normalized by maximum and minimum values due to dataset-specific invalid regions.

\subsection{Unbalanced Timestep Sampling Strategy}
When applying the unbalanced timestep sampling strategy, the timesteps, \ie, $t_x$ and $t_y$, are separately sampled from the timestep distributions $f(t)$ and $g(t)$, respectively, or vice versa. This is applied with a 50\% probability during training, while for the remaining 50\%, the same timestep sampled from 
$f(t)$ is used for both $t_x$ and $t_y$. The timestep distribution is as follows:
\begin{equation}
    f(t) = 1 - \frac{\sigma(z) \cdot s}{1 + (s - 1) \cdot \sigma(z)}, \quad \text{where } z \sim \mathcal{N}(0,1).
\end{equation}

The $\sigma(\cdot)$ denotes the sigmoid function. In $f(t)$, which is suggested by our base training code\footnote{https://github.com/kohya-ss/sd-scripts/tree/sd3}, $s$ is set to 3.1582. We set $s$ to 0.25 to obtain $g(t)$.

\subsection{Architecture of JointDiT}
To build the depth branch, we add LoRAs~\cite{hu2022lora} to the original Flux architecture~\cite{Flux2024}. Specifically, we add LoRAs to the components connected before and after the attention mechanisms of the multi-modal diffusion transformer (MM-DiT) and parallel diffusion transformer (P-DiT) blocks~\cite{esser2403scaling,dehghani2023scaling} that constitute Flux. Table~\ref{tab:lora_flux} summarizes the LoRA-applied components in the MM-DiT and P-DiT blocks. We use a LoRA rank of 64 for both MM-DiT and P-DiT, and apply relatively larger ranks of 512 or 1024 to the input stage. The alpha value is set to half of the corresponding rank. 

To design the joint connection module, we adopt the joint cross-attention module from UniCon~\cite{li2024simple}, followed by a zero-initialized linear projection layer. The adaptive scheduling weight is applied subsequently.

\section{Additional Experiments}

\subsection{Advantages of Joint RGB-Depth Modeling}
As mentioned in the main paper, we observe that joint RGB-Depth generation tends to yield more plausible 3D lifting results compared to estimating depth from generated images. Figure~\ref{figure:realistic} presents the 3D lifting results by showing top and side views. When using the depth generated by our JointDiT, the results exhibit more well-structured and volumetric 3D geometry than those produced by Marigold~\cite{ke2024repurposing} and Depth-Anything-V2~\cite{yang2025depth}.

Furthermore, as also discussed in the main paper, our joint generation approach enables plausible depth synthesis even in illustration domains, where depth estimation methods often struggle. Additional qualitative results are presented in Figure~\ref{figure:illustration}.

\subsection{Jointly Generated Image Quality}

\begin{table*}[t]
    \centering
    \begin{tabular}{lccccccccc}
        \toprule
        \multirow{3}{*}{Method} & \multicolumn{3}{c}{ImageNet 6K} & \multicolumn{3}{c}{Pexels 6K} & \multicolumn{3}{c}{MSCOCO 30K} \\
        \cmidrule(lr){2-4}\cmidrule(lr){5-7}\cmidrule(lr){8-10}
        & \multicolumn{3}{c}
        {ImageReward} & \multicolumn{3}{c}{ImageReward} & 
        \multicolumn{3}{c}{ImageReward} \\
        \cmidrule(lr){2-4}\cmidrule(lr){5-7}\cmidrule(lr){8-10}
        & Rank1$\uparrow$ & Rank2 & Rank3$\downarrow$ & Rank1$\uparrow$ & Rank2 & Rank3$\downarrow$ & Rank1$\uparrow$ & Rank2 & Rank3$\downarrow$ \\
        \midrule
        LDM3D~\cite{stan2023ldm3d} & 27.56 & 35.90 & 36.54 &	26.21 &	33.42 &	40.37 &	27.74 &	34.04 &	38.22 \\
        JointNet~\cite{zhang2023jointnet} & 29.91 &	33.32 &	36.77 &	31.65 &	33.87 &	34.48 &	28.85 &	35.70 &	35.46 \\
        \midrule
        Ours & \textbf{42.53} &	30.79 &	\textbf{26.69} & \textbf{42.14} &	32.72 &	\textbf{25.15} &	\textbf{43.41} &	30.26 &	\textbf{26.32} \\
        \bottomrule
    \end{tabular}
    \caption{\textbf{Human preference evaluation on images jointly generated by joint generation methods~\cite{stan2023ldm3d,zhang2023jointnet} and Ours.} We assess the human preference using ImageReward~\cite{xu2023imagereward} that was trained to estimate human preference. With both joint generation models and ours, we conduct joint generation using the same text prompts and rank the results with ImageReward, obtaining the percentage for each ranking. Our JointDiT achieved the highest rank 1 percentage and the lowest rank 3 percentage across all datasets.
    }        
    \label{tab:humanpreference}
\end{table*}

\begin{figure}[t]
\centering
\includegraphics[width=1\linewidth]{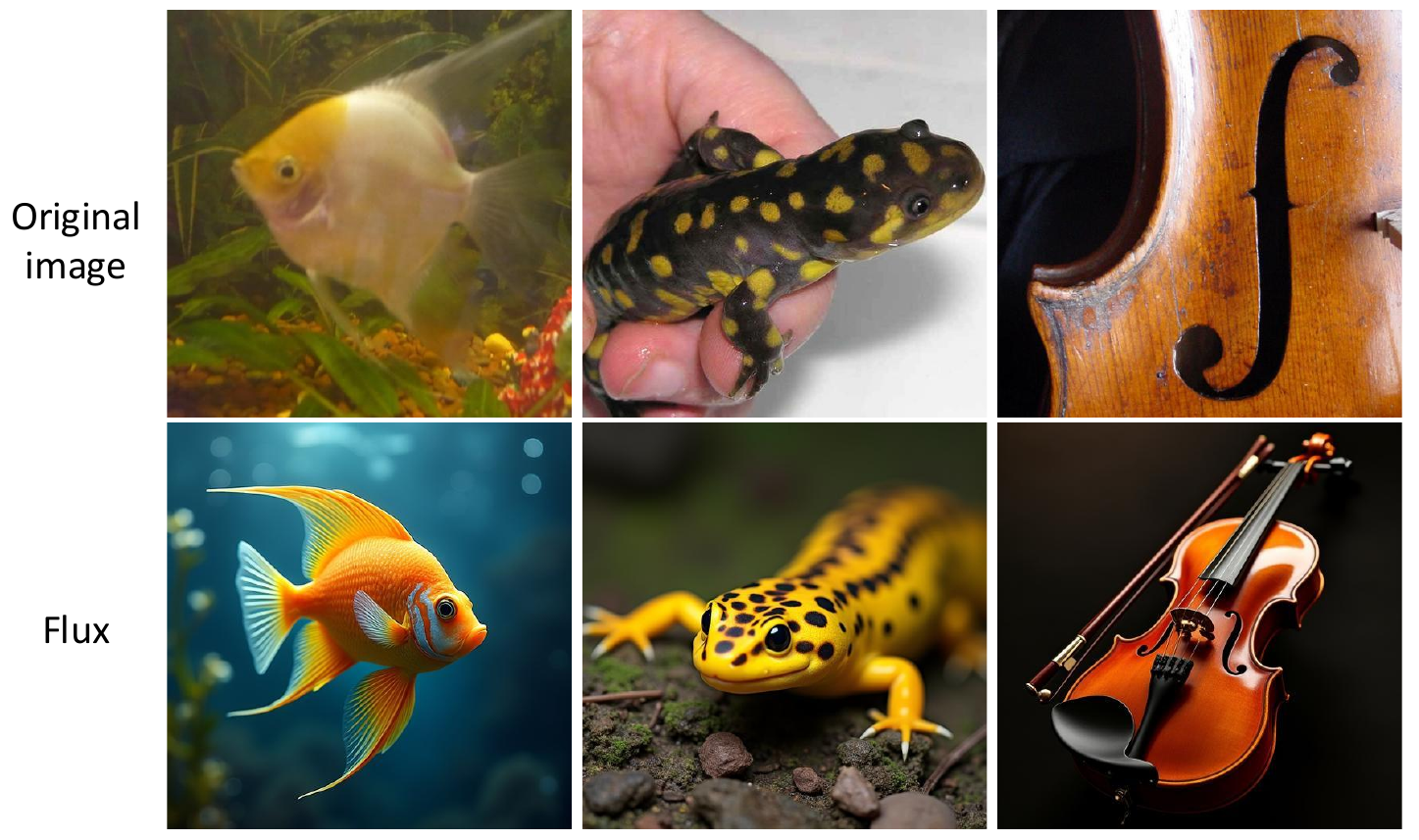}
    \caption{
    \textbf{Comparison between original images and images generated by Flux~\cite{Flux2024} on the ImageNet~\cite{russakovsky2015imagenet} 6K dataset.} Flux often generates stylized images, which leads to a higher FID between the real image dataset and the generated images.
    }  \label{figure:flux_samples}
\end{figure}

We quantitatively compare the quality of jointly generated images from JointNet~\cite{zhang2023jointnet}, LDM3D~\cite{stan2023ldm3d}, and our method.
For evaluation, we use the dataset from Section 4.4 of the main paper, \ie, ImageNet 6K, Pexels 6K, and MSCOCO 30K.
We measure the Inception Score (IS)~\cite{salimans2016improved}, Fréchet Inception Distance (FID)~\cite{heusel2017gans}, and CLIP similarity~\cite{radford2021learning} as our evaluation metrics. Table~\ref{tab:quan_joint} summarizes the results. We also include the results of baseline diffusion models, \ie, Stable diffusion~\cite{rombach2022high} and Flux~\cite{Flux2024}. 
Interestingly, Flux achieves relatively high FID scores across all evaluation datasets despite its outstanding text-to-image generation capability. We observe that Flux often generates stylized images. Figure~\ref{figure:flux_samples} shows samples from ImageNet 6K and the corresponding images generated by Flux. The generated samples appear surreal, which leads to a higher FID between them and the real image dataset. Our model achieves a lower FID score than Flux by learning the joint distribution of images and their corresponding depth maps on the real dataset. However, our IS score is lower than that of Flux, likely due to the limited size of the training dataset.

Among the joint generation models,  LDM3D shows relatively poor performance. Our method achieves comparable performance to JointNet. To further assess image generation quality, we evaluate the human preference score using ImageReward~\cite{xu2023imagereward}, a trained model that estimates human preference for given text prompts and images. We measure the human preference ranking of the images generated by the joint generation model from the same text prompt. Table~\ref{tab:humanpreference} summarizes the percentage of each method on each evaluation dataset. Our method shows the highest rank 1 percentage and the lowest rank 3 percentage across all evaluation datasets. Compared to LDM3D, JointNet achieves moderately better performance.

\subsection{Ablation of the LoRA's Rank}
We adopt a LoRA rank of 64 in the DiT blocks of our JointDiT model. To analyze the effect of the LoRA rank, we train our model with different LoRA ranks and evaluate depth estimation performance on the NYUv2 and ScanNet datasets~\cite{silberman2012indoor,dai2017scannet}. As shown in Table~\ref{tab:ablation_lora}, 
as the LoRA rank increases, the depth estimation performance improves, achieving the best performance at the LoRA rank of 64. We did not increase the LoRA rank beyond 64 because the number of model parameters grows exponentially.

\subsection{Analysis of Failure Cases}
\begin{figure}[t]
\centering
\includegraphics[width=\linewidth]{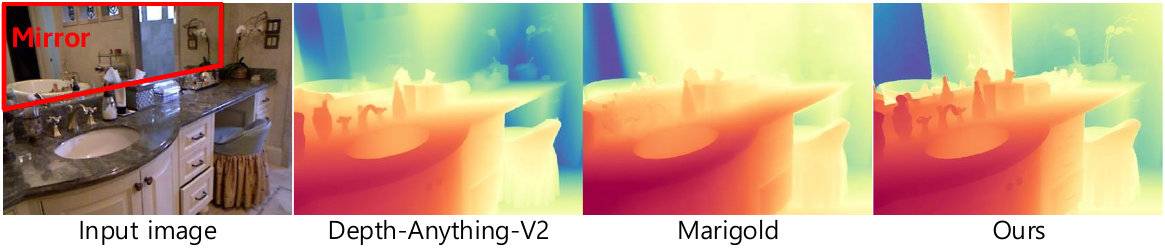}
    \vspace{-0.2cm}
    \caption{
    \textbf{Failure cases in depth estimation.} \textcolor{red}{Red} and \textcolor{Blue}{Blue} areas indicate near and far depth predictions, respectively.
    }     
    \vspace{-0.3cm}
    \label{figure:failure}
\end{figure}

We observe that our method shares similar limitations with depth estimation methods~\cite{yang2025depth,ke2024repurposing}, particularly in handling reflective surfaces such as mirrors. As shown in Fig.~\ref{figure:failure}, both our model and depth estimation models fail to recognize mirrors as flat and planar regions.

\subsection{Joint Panorama Generation} 
\begin{figure*}[t]
\centering
\includegraphics[width=0.96\linewidth]{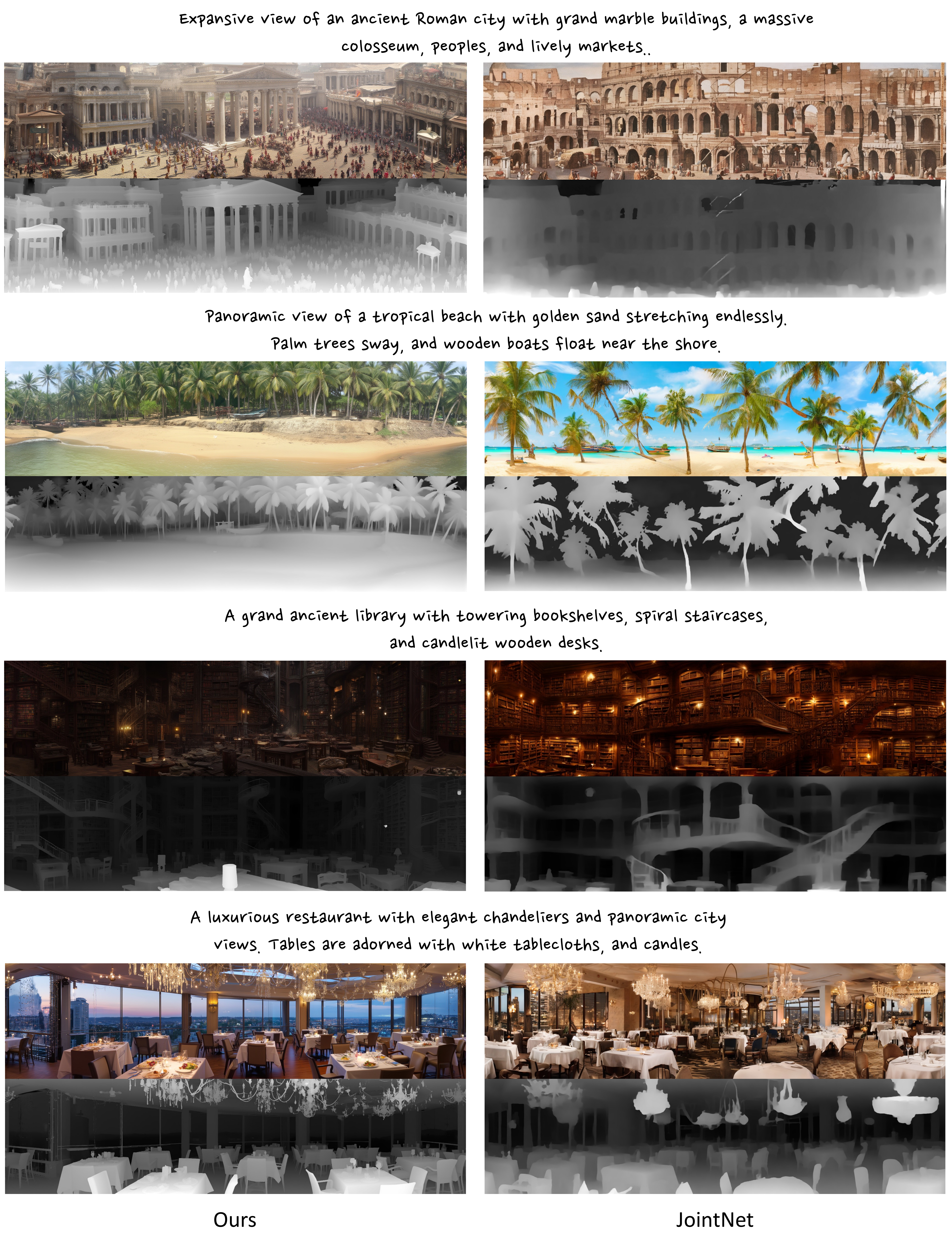}
    \caption{
    \textbf{RGB-D panoramic generation results of JointNet and Ours.} Our JointDiT generates more three-dimensional and sharper images and depth maps compared to JointNet.
    } 
    \label{figure:supple_panorama}
\end{figure*}

JointDiT can be used for RGB-D panorama generation as well.
For panorama generation, we strictly follow the JointNet~\cite{zhang2023jointnet} method combining whole and tile-based denoising strategies~\cite{bar2023multidiffusion,jimenez2023mixture}, to ensure a fair comparison. We denoise image and depth tiles by using joint generative diffusion models. During only early steps, we perform denoising on the entire panorama, and throughout all steps, we aggregate model estimations from both overlapped individual tiles and the whole panorama. Figure~\ref{figure:supple_panorama} demonstrates the RGB-D panorama results. Compared to JointNet, JointDiT shows clear and structurally reasonable images along with sharp depth maps.

\section{Additional Qualitative Results}
In this section, we demonstrate diverse qualitative results on depth estimation and depth-conditioned image generation.

\begin{table}[t]
    \centering
    \begin{tabular}{ccccc}
        \toprule
        \multirow{2}{*}{LoRA rank} & \multicolumn{2}{c}{NYUv2~\cite{silberman2012indoor}} & \multicolumn{2}{c}{ScanNet~\cite{dai2017scannet}} \\
        \cmidrule(l){2-3}\cmidrule(l){4-5}
        &  AbsRel $\downarrow$ & $\delta_1$ $\uparrow$ &  AbsRel $\downarrow$ & $\delta_1$ $\uparrow$ \\
        \midrule
        16 & 9.1 & 90.6 & 9.8 & 89.7\\
        32 & 6.6 & 95.7 & 8.5 & 92.4 \\
        \midrule
        64 (Ours) & \textbf{5.7} & \textbf{96.9} & \textbf{6.6} & \textbf{95.7} \\
        \bottomrule
    \end{tabular}
    \caption{\textbf{Ablation studies of the rank of LoRA.} We evaluate the depth estimation performance on NYUv2 and ScanNet while varying the LoRA rank. The results show that performance improves as the LoRA rank increases.
    }
    \label{tab:ablation_lora}
\end{table}

\begin{figure*}[t]
\centering
\includegraphics[width=0.88\linewidth]{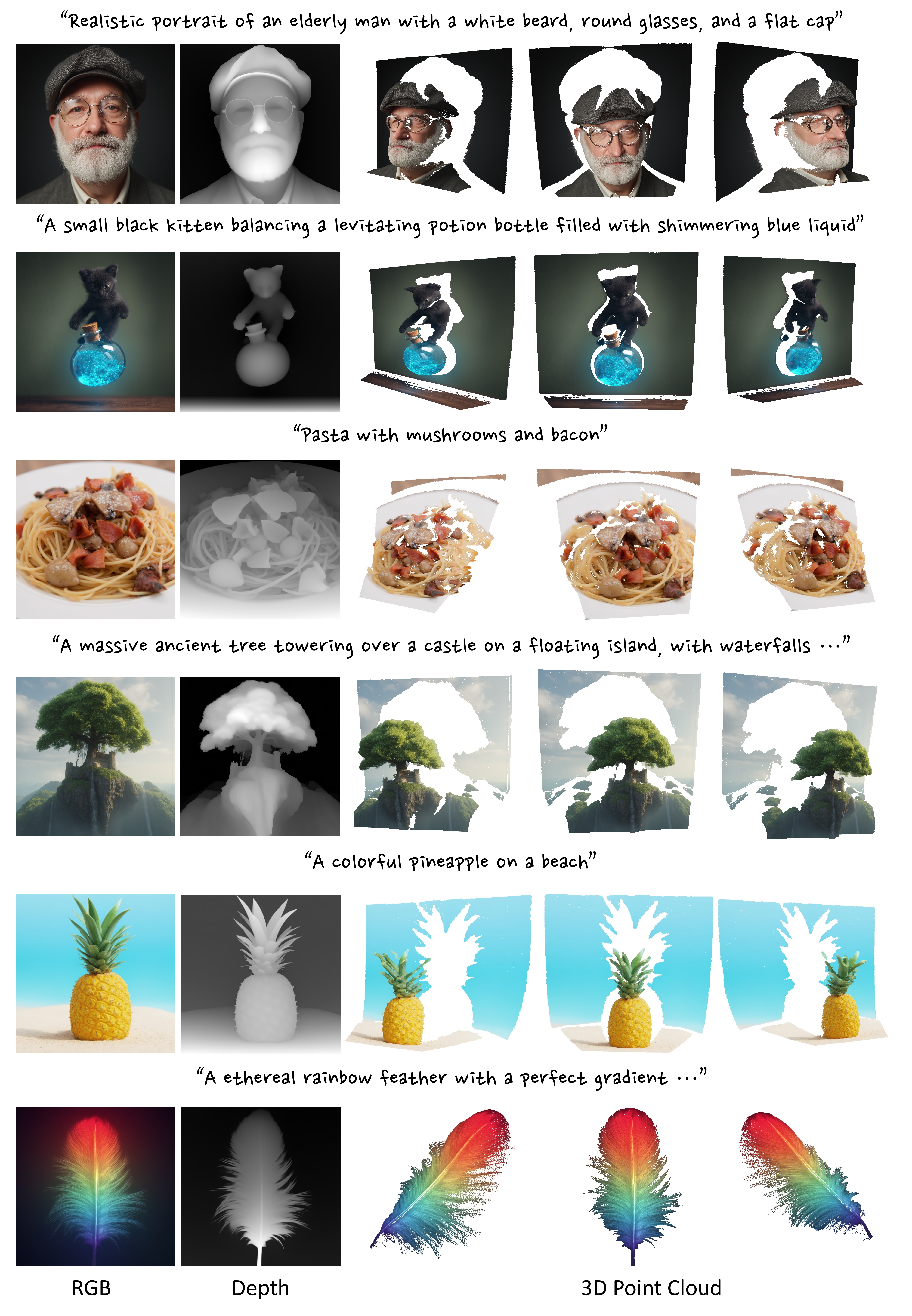}
    \vspace{-3mm}
    \caption{
    \textbf{Joint generation results of JointDiT.} The joint generated images and depths are geometrically reasonable in 3D.
    } 
    \label{figure:supple_joint_ours}
\end{figure*}

\begin{figure*}[t]
\centering
\includegraphics[width=0.96\linewidth]{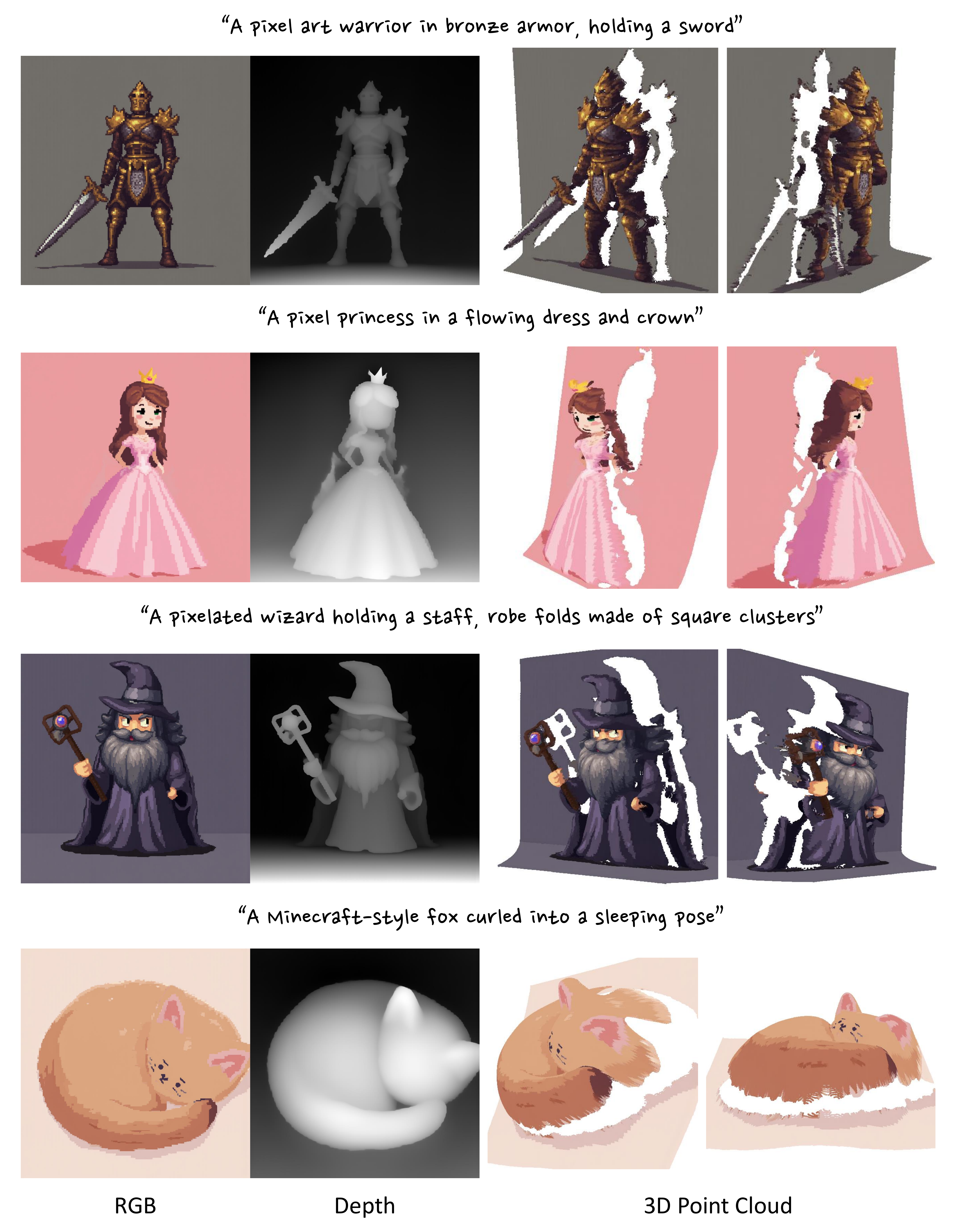}
    \caption{
    \textbf{Joint generation results in illustration domains.} The jointly generated images and depths from JointDiT produce geometrically plausible 3D structures, even in stylized domains.
    } 
    \label{figure:illustration}
\end{figure*}

\paragraph{Joint generation} Utilizing our JointDiT model, we generate images and corresponding depth maps. We visualize the images and depths with their 3D lifting results. As shown in Fig.~\ref{figure:supple_joint_ours}, our joint generation results are geometrically reasonable in 3D, with the surface characteristics of the images being well-preserved in the 3D space (e.g., smooth or rough textures). Furthermore, Figure~\ref{figure:illustration} highlights the effectiveness of our joint generation approach in illustration domains, where plausible 3D structures are obtained despite the inherent difficulty of estimating geometry from stylized images.

\begin{figure*}[t]
\centering
\includegraphics[width=0.96\linewidth]{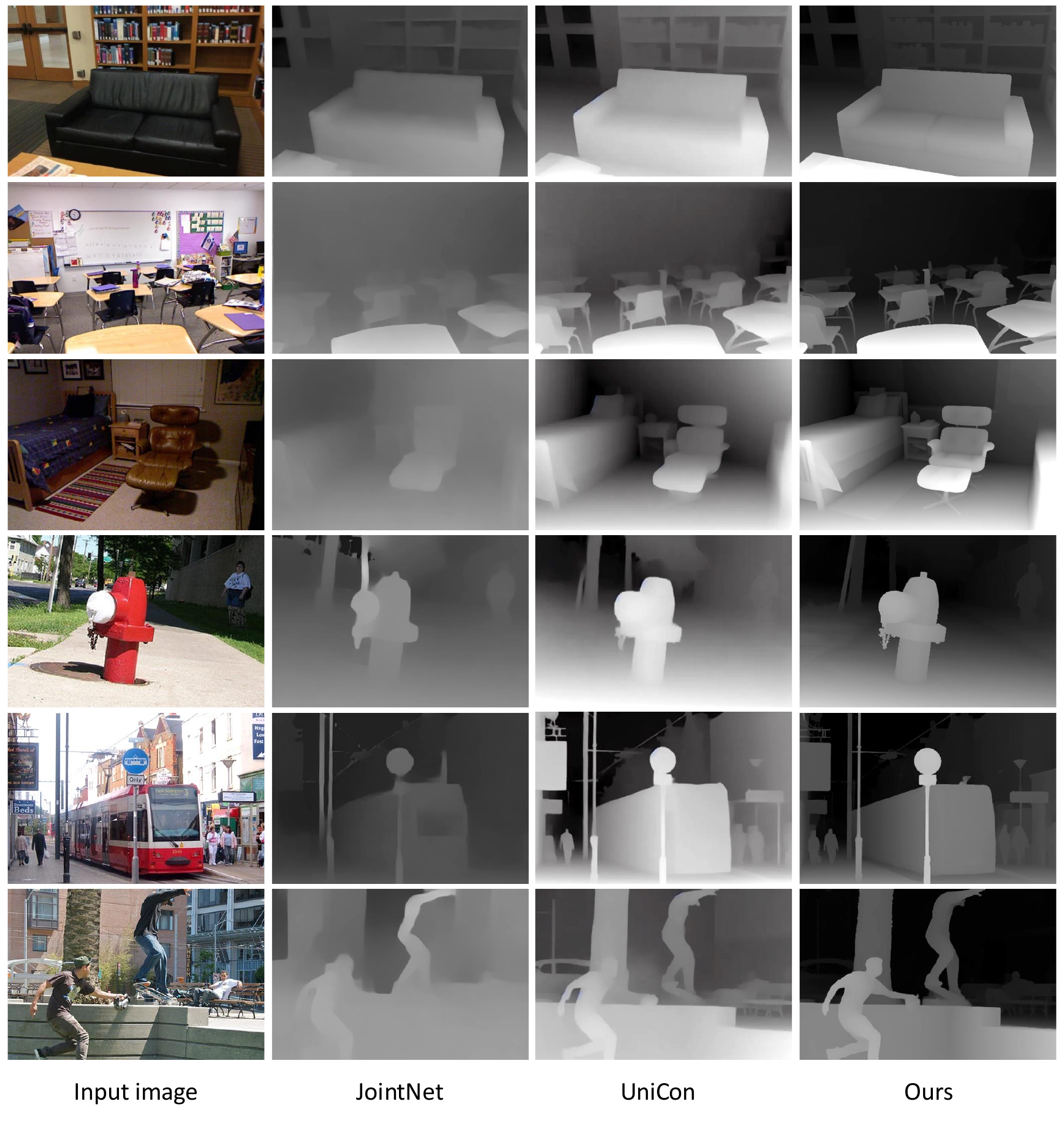}
    \caption{
    \textbf{Depth estimation results of joint generation models.} We visualize the depth estimation results of JointNet, UniCon, and our method on the NYUv2, ScanNet, MSCOCO dataset~\cite{dai2017scannet,silberman2012indoor,lin2014microsoft}. Our method shows sharp and fine-detailed depth visualization, which aligns with the trends observed in the qualitative results.
    } 
    \label{figure:supple_depth}
\end{figure*}

\paragraph{Depth estimation} We visualize the depth estimation results of joint generation methods that support depth estimation, \ie, JointNet~\cite{zhang2023jointnet}, UniCon~\cite{li2024simple}, and Ours. We obtain the depth estimation results from the publicly available code. Specifically, while UniCon does not provide raw depth through its Gradio demo, we can obtain depth estimation visualization results. To estimate depth, we provide each model with empty text prompts.
To demonstrate the results across various scenarios, we acquire depth maps estimated from the NYUv2, ScanNet, and MSCOCO datasets~\cite{silberman2012indoor,dai2017scannet,lin2014microsoft}.  
Figure~\ref{figure:supple_depth} illustrates the results. Compared to JointNet and UniCon, our method captures fine details in the depth and the shape of thin objects. This aligns with the trends observed in the quantitative results.

\begin{figure*}[t]
\centering
\includegraphics[width=0.96\linewidth]{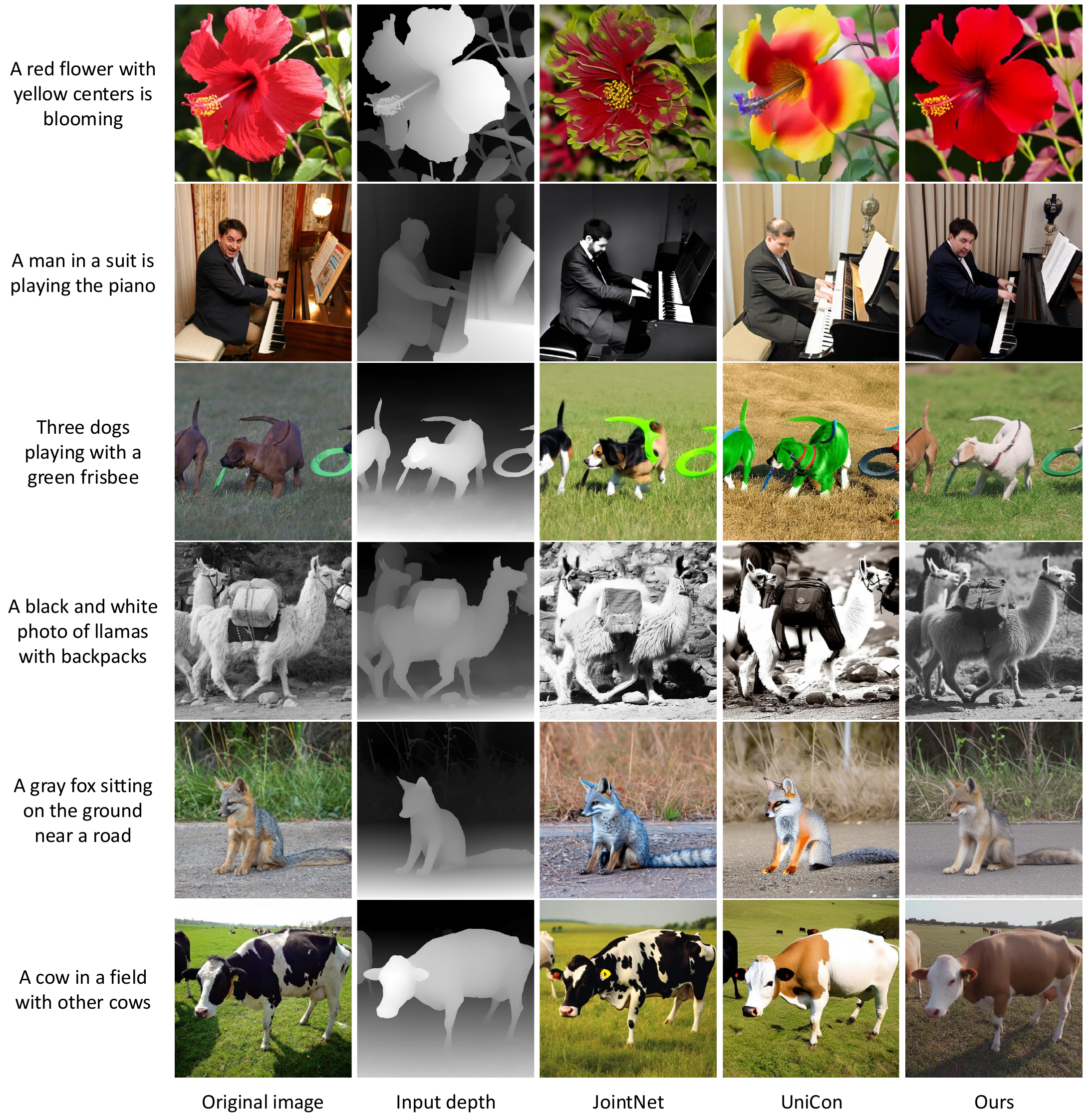}
    \caption{
    \textbf{Depth-conditioned image generation results of JointNet, UniCon, and Ours.}
     JointNet and UniCon often fail to reflect the text prompt properly, \eg, the green dog generated by UniCon and the flower with green petals generated by JointNet. Our JointDiT generates images that better reflect the text prompt and depth map, producing more realistic results compared to other methods.
     }
     \label{figure:supple_depth_to_img}
\end{figure*}

\paragraph{Depth-conditioned image generation} We visualize the depth-conditioned image generation results of JointNet, UniCon, and our method. We utilize publicly available code for the other two methods. To generate the results, we obtain depth maps and text prompts from ImageNet 6K using Depth-Anything-V2~\cite{yang2025depth} and LLaVA~\cite{liu2023visual}. For JointNet, we provide the depth estimation from MiDaS~\cite{ranftl2020towards}, as it was trained using MiDaS' depth estimation. Figure~\ref{figure:supple_depth_to_img} demonstrates the results. JointNet and UniCon generally generate images that match the given depth and text prompts, but they sometimes do not fully understand the text prompt. For example, UniCon generated a green dog instead of a green frisbee, and JointNet failed to fully generate a red flower. In comparison, our JointDiT shows generation results that are well aligned with the given depth and text prompts, and we observe that it generates more realistic images than the other models.

\end{document}